\def\paperlanguage{}
\pdfoutput=1 

\documentclass[letterpaper, 10 pt, journal, twoside]{IEEEtran}  

\usepackage{bm}
\usepackage{cite}
\include{preamble}

\newcommand{\ctext}[1]{\raise0.2ex\hbox{\textcircled{\scriptsize{#1}}}}
\definecolor{bvqa}{rgb}{0.7,0.7,0}
\definecolor{mvqa}{rgb}{0.7,0,0}
\definecolor{itr}{rgb}{0.7,0,0.7}
\definecolor{vg}{rgb}{0,0,0.7}
\definecolor{dic}{rgb}{0,0.7,0.7}

\IEEEoverridecommandlockouts                              

\title{
  {
    \switchlanguage%
    {%
      Characteristics, Management, and Utilization of\\Muscles in Musculoskeletal Humanoids:\\Empirical Study on Kengoro and Musashi
    }%
    {%
      筋骨格ヒューマノイドの筋肉の特徴, 対処, 応用:\\KengoroとMusashiを例に
    }%
  }
}

\author{Kento Kawaharazuka$^{*1}$, Kei Okada$^{1}$, and Masayuki Inaba$^{1}$
  \thanks{$^{*}$ Corresponding Author: Kento Kawaharazuka}
  \thanks{$^{1}$ The authors are with the Department of Mechano-Informatics, Graduate School of Information Science and Technology, The University of Tokyo, 7-3-1 Hongo, Bunkyo-ku, Tokyo, 113-8656, Japan.
    {\texttt\small [kawaharazuka, k-okada, inaba]@jsk.t.u-tokyo.ac.jp}
  }
}
\begin{document}

\maketitle
\thispagestyle{empty}
\pagestyle{empty}

\begin{abstract}
  \switchlanguage%
  {%
    Various musculoskeletal humanoids have been developed so far, and numerous studies on control mechanisms have been conducted to leverage the advantages of their biomimetic bodies.
    However, there has not been sufficient and unified discussion on the diverse properties inherent in these musculoskeletal structures, nor on how to manage and utilize them.
    Therefore, this study categorizes and analyzes the characteristics of muscles, as well as their management and utilization methods, based on the various research conducted on the musculoskeletal humanoids we have developed, Kengoro and Musashi.
    We classify the features of the musculoskeletal structure into five properties: Redundancy, Independency, Anisotropy, Variable Moment Arm, and Nonlinear Elasticity.
    We then organize the diverse advantages and disadvantages of musculoskeletal humanoids that arise from the combination of these properties.
    In particular, we discuss body schema learning and reflex control, along with muscle grouping and body schema adaptation.
    Also, we describe the implementation of movements through an integrated system and discuss future challenges and prospects.
  }%
  {%
    これまで様々な筋骨格ヒューマノイドが作られ, その生物模倣型の身体の利点を活かす様々な制御に関する研究が行われてきた.
    一方で, これら筋骨格構造に潜む様々な性質, そしてそれらをどのように管理し, どのように利用するかについては, まだ十分に統一的な議論がなされていない.
    そこで本研究では, これまで我々が開発してきた筋骨格ヒューマノイドKengoroとMusashiを例に, 様々な実験から得られた筋肉の特徴と, その管理, 応用方法について分類・考察する.
    我々は筋骨格構造の特徴を冗長性・独立性・異方性・可変モーメンタム・非線形弾性という5つの特性に分類し, これらの組み合わせから生まれる筋骨格ヒューマノイドの多様な利点と欠点について整理する.
    特に, 身体図式学習と反射制御, それに付随する筋のグルーピングと身体変化への適応について述べる.
    また, それらを統合した全体システムによる動作実現, 今後の課題と展望について述べる.
  }%
\end{abstract}


\section{Introduction}\label{sec:introduction}
\switchlanguage%
{%
  Various musculoskeletal humanoids have been developed so far \cite{jantsch2013anthrob, gravato2010ecce1, nakanishi2012kenshiro, asano2016kengoro, kawaharazuka2019musashi}.
 These have been researched for diverse purposes, such as achieving a constructive understanding of the human body, validating the advantages of biomimetic structures, and transferring the benefits of imitating human anatomy to robotic technologies.

  The musculoskeletal structure includes bones, joints, ligaments, muscles, and skin, among which muscles play a particularly important role.
  Muscles possess diverse characteristics not found in general axis-driven robots \cite{hirai1998asimo, kaneko2004hrp2}, creating not only advantages but also disadvantages for musculoskeletal humanoids.
  For example, redundant muscles and their nonlinear elasticity \cite{calvo2010passive} enable variable stiffness control \cite{kobayashi1998tendon}.
  On the other hand, the redundancy introduced by numerous muscles makes modeling and control more challenging \cite{kawaharazuka2018online}, increasing the risk of internal forces accumulating and causing damage due to modeling errors.
  Thus, the properties of muscles are highly diverse and complex, and the advantages and disadvantages that arise from their combinations are even more intricate.

  To date, numerous studies have explored the characteristics, advantages, and disadvantages of musculoskeletal systems.
  In particular, much attention has been given to redundancy.
  Research topics include variable stiffness utilizing redundancy and nonlinear elasticity \cite{stanev2019stiffness, tsuboi2022endpoint}, the relationship between redundancy and muscle synergies \cite{sharif2019synergies}, the impact of redundancy on contact force prediction \cite{moissenet2016influence}, approaches to addressing muscle rupture using redundancy \cite{almanzor2024utilising}, as well as studies that view redundancy as a drawback and explore control methods to reduce it \cite{zhong2019reducing}.
  Additionally, studies cover a wide range of topics, such as the contribution of biarticular muscles to balance control and energy efficiency \cite{sharbafi2016biarticular, nejadfard2020biarticular}, and learning the complex relationships between muscles and joints that are difficult to model \cite{marjaninejad2019tendon, hagen2021insideout}.
  However, most of these studies are conducted in simulations or on simplified 2D models, with very few focusing on actual full-body musculoskeletal robots.
  Furthermore, there is still a lack of unified discussion on the characteristics of muscles in robots, how these characteristics should be managed, and how they can be effectively utilized.
  Most existing discussions remain fragmented and isolated.
}%
{%
  これまで様々な筋骨格ヒューマノイドが開発されてきた\cite{jantsch2013anthrob, gravato2010ecce1, nakanishi2012kenshiro, asano2016kengoro, kawaharazuka2019musashi}.
  それらは, 人間の身体の構成論的理解や生物の身体の利点検証, 人体模倣構造の利点のロボットへの技術転用など, 多様な目的を持って研究されてきている.

  筋骨格構造には骨や関節, 靭帯や筋肉, 皮膚などが含まれるが, その中でも筋肉は特に重要な役割を担っている.
  この筋肉には一般的な軸駆動型ロボット\cite{hirai1998asimo, kaneko2004hrp2}にはない多様な特性があり, これらは筋骨格ヒューマノイドにおいて利点だけでなく, 欠点も生み出す.
  例えば, 冗長な筋肉とそれらが持つ非線形弾性\cite{calvo2010passive}により可変剛性制御が可能な利点がある\cite{kobayashi1998tendon}.
  一方で, 多数の筋肉によって冗長性が増し, モデル化や制御がより困難になり\cite{kawaharazuka2018online}, モデル化誤差によって内力がたまり破損してしまう危険性がある.
  このように, 筋肉が持つ特性は非常に多様かつ複雑であり, それら特性が組み合わさって引き起こす利点や欠点はより複雑となる.

  これまで, 筋骨格身体の特性とその利点・欠点については様々な研究が行われてきた.
  特に冗長性に関する研究が多く, 冗長性を利用した可変剛性に関する研究\cite{stanev2019stiffness, tsuboi2022endpoint}, 冗長性と筋シナジーの関係性に関する研究\cite{sharif2019synergies}, 冗長性が接触力予測に及ぼす影響に関する研究\cite{moissenet2016influence}, 冗長性を用いた筋破断に対する対処方法の研究\cite{almanzor2024utilising}, 冗長性を欠点としてそれを削減する制御方法の研究\cite{zhong2019reducing}などが行われてきた.
  この他にも, 二関節筋のバランス制御やエネルギー効率への寄与に関する研究\cite{sharbafi2016biarticular, nejadfard2020biarticular}やモデル化の難しい筋肉と関節の関係性を学習する研究\cite{marjaninejad2019tendon, hagen2021insideout}など, 研究のバラエティは多岐にわたる.
  しかし, これらはシミュレーション上やシンプルな2次元平面上のシンプルなモデルでの研究がほとんどであり, 全身の筋骨格ロボット実機に関する研究は非常に少ない.
  また, 現状ロボットにおける筋肉にはどのような特性があるか, これらをどのように管理し, どのように利用するかについては, 個別の議論ばかりでまだ十分に統一的な議論がなされていない.
}%

\begin{figure}[t]
  \centering
  \includegraphics[width=0.95\columnwidth]{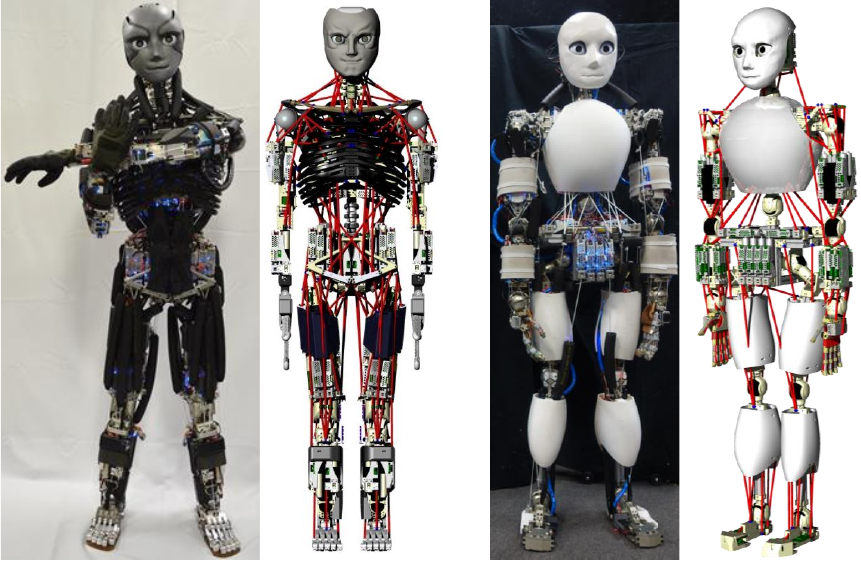}
  \vspace{-1.0ex}
  \caption{The developed musculoskeletal humanoids: Kengoro and Musashi}
  \label{figure:robots}
  \vspace{-2.0ex}
\end{figure}

\switchlanguage%
{%
  In this study, we empirically categorize and analyze the characteristics of muscles, as well as their management and utilization methods based on the various research conducted on the musculoskeletal humanoids we have developed, Kengoro \cite{asano2016kengoro} and Musashi \cite{kawaharazuka2019musashi} shown in \figref{figure:robots}.
  We classify the characteristics of the musculoskeletal structure into five properties: Redundancy, Independency, Anisotropy, Variable Moment Arm, and Nonlinear Elasticity, organizing the diverse advantages and disadvantages of musculoskeletal humanoids that arise from these combinations.
  In particular, we describe how to manage and utilize muscle characteristics concerning body schema learning, reflex control, muscle grouping, and body schema adaptation.
  Additionally, we discuss the implementation of movements through an integrated whole-body system, summarizing future challenges and perspectives.

  This study focuses on musculoskeletal humanoids that do not use pulleys at the joints, where the moment arm from the joints to the muscles and the muscle Jacobian are not constant, creating a more human-like musculoskeletal structure.
  Therefore, wire-driven robots that use pulleys at the joints to maintain constant moment arm \cite{endo2019superdragon, koganezawa1999stiffness} are not directly addressed.
  However, these are simpler systems than musculoskeletal humanoids, so similar theories can be applied.
  Also, instead of musculoskeletal humanoids operated by pneumatic artificial muscles \cite{ogawa2011pneumatbs, mizuuch2012buenwa}, this study focuses on those that use electric motors to wind wires.
  Most properties are the same for pneumatic types and similar theories can be applied, but for the sake of unified explanation, we focus on the structure where wires are wound by electric motors.
  Additionally, note that the hardware is fixed and this study focuses on the software.

  The structure of this study is as follows.
  In \secref{sec:overview}, we provide an overview of the characteristics of musculoskeletal humanoids and their advantages and disadvantages.
  In \secref{sec:detail}, we describe how to address these advantages and disadvantages through body schema learning, reflex control, muscle grouping, and body schema adaptation in the musculoskeletal humanoids Kengoro and Musashi.
  In \secref{sec:experiments}, we describe several motion experiments conducted by integrating these software systems.
  Finally, in \secref{sec:discussion}, we discuss future challenges and perspectives based on the research so far.
}%
{%
  そこで本研究では, これまで我々が開発してきた筋骨格ヒューマノイドKengoro \cite{asano2016kengoro}とMusashi \cite{kawaharazuka2019musashi}を例に(\figref{figure:robots}), 筋肉の特徴と, その管理, 応用方法について分類・考察する.
  我々は筋骨格構造の特性を冗長性・独立性・異方性・可変モーメンタム・非線形弾性という5つの特性に分類し, これらの組み合わせから生まれる筋骨格ヒューマノイドの多様な利点と欠点について整理する.
  特に, 身体図式学習と反射制御, それに付随する筋のグルーピングと身体変化への適応について, どのように筋肉の特性を管理, 利用可能かを述べる.
  また, それらを統合した全体システムによる動作実現について述べ, 今後の課題と展望についてまとめる.

  なお, 本研究は筋骨格ヒューマノイドの中でも, 関節にプーリを使わず, 関節から筋へのモーメンタム, 筋長ヤコビアンが一定ではない, より人間らしい筋骨格ヒューマノイドに焦点を当てる.
  そのため, 関節にプーリを用いて筋のモーメンタムを一定にしたワイヤ駆動ロボット\cite{endo2019superdragon, koganezawa1999stiffness}は直接扱わない.
  一方で, これらは筋骨格ヒューマノイドよりも簡易な系なため, 同様の理論を適用可能である.
  また, 空気圧人工筋肉により動作する筋骨格ヒューマノイド\cite{ogawa2011pneumatbs, mizuuch2012buenwa}ではなく, モータでワイヤを巻き取るタイプの筋骨格ヒューマノイドに絞って説明を進める.
  もちろん空気圧タイプでもほとんどの性質は同じであり同様の理論が適用可能であるが, 説明を統一するため, モータによりワイヤを巻き取る構造を中心に扱っている.
  加えて, ハードウェアについては固定し, 体内ソフトウェアにおける制御の話に焦点を当てている点にも注意されたい.

  本研究の構成は以下の通りである.
  まず, \secref{sec:overview}では, 筋骨格ヒューマノイドの特性とその利点/欠点について概要を述べる.
  次に, \secref{sec:detail}では, それら利点/欠点を対処する, 筋骨格ヒューマノイドKengoroとMusashiの身体図式学習, 反射制御, 付随する筋のグルーピング, 身体変化への適応について述べる.
  次に, \secref{sec:experiments}では, これらソフトウェアを統合したシステムによるいくつかの動作実験について述べる.
  最後に, \secref{sec:discussion}では, これまでの研究を踏まえて, 今後の課題と展望について述べる.
}%

\begin{figure}[t]
  \centering
  \includegraphics[width=0.85\columnwidth]{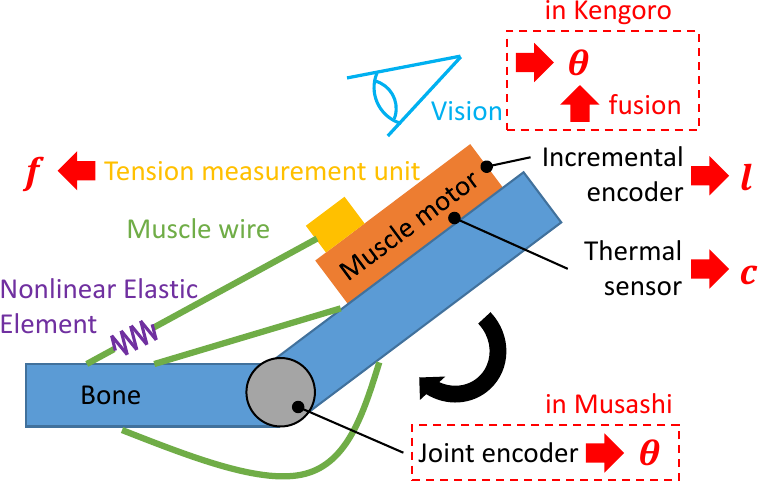}
  \vspace{-1.0ex}
  \caption{The basic musculoskeletal structure: the components include bones, joints, muscle actuators, wires, and nonlinear elastic elements. Muscle length $l$, muscle tension $f$, muscle temperature $c$, and joint angle $\theta$ (directly or indirectly depending on the robot) can be measured.}
  \label{figure:structure}
  \vspace{-2.0ex}
\end{figure}

\section{Overview of Characteristics, Management, and Utilization of Muscles in Musculoskeletal Humanoids}\label{sec:overview}

\subsection{Basic Musculoskeletal Structure}\label{subsec:structure}
\switchlanguage%
{%
  First, the basic structure of the musculoskeletal humanoid handled in this study is described.
  As shown in \figref{figure:structure}, the musculoskeletal structure consists of bones, joints, muscle actuators, wires, and nonlinear elastic elements.
  The muscle actuator drives the joint by winding the wire using a pulley attached to the motor.
  A nonlinear elastic element is attached to the end of the wire to mimic the nonlinear elasticity of muscles.
  Wires are often made of chemical fibers like Dyneema, which are resistant to friction.
  The muscle actuators are equipped with incremental encoders, tension measurement units, and temperature sensors, allowing the measurement of muscle length $l$, muscle tension $f$, and muscle temperature $c$.
  Here, the robot's joint angles are first returned to their initial position, where the muscle lengths are set to zero, and the encoder values are initialized.
  Some robots (e.g. Musashi \cite{kawaharazuka2019musashi}) have encoders attached to the joints, allowing for the measurement of the joint angle $\theta$.
  However, when imitating complex joints such as human ball joints and the scapula (e.g. Kengoro \cite{asano2016kengoro}), joint angle sensors cannot be used.
  In such cases, if the position of the end-effector can be measured using vision sensors, the joint angle $\theta$ can be estimated from the relationship between the change in muscle length $l$ and the position of the end-effector \cite{kawaharazuka2018online}.
  Muscles are redundantly arranged around the joints, and there are muscles that span not only a single joint but also multiple joints.

  Next, the fundamental equations that describe the relationship between joints and muscles in musculoskeletal humanoids are presented.
  Let the number of muscles be $M$, the number of joints be $N$, muscle tension be $\bm{f}$, joint torque be $\bm{\tau}$, end-effector force be $\bm{F}$, muscle length be $\bm{l}$, joint angle be $\bm{\theta}$, and end-effector position be $\bm{x}$.
  The following equations generally hold,
  \begin{align}
    \bm{l} &= \bm{g}_{m}(\bm{\theta}) \\
    \dot{\bm{l}} &= \bm{G}(\bm{\theta})\dot{\bm{\theta}} \\
    \bm{\tau} &= -\bm{G}^{T}(\bm{\theta})\bm{f} \\
    \bm{x} &= \bm{g}_{j}(\bm{\theta}) \\
    \dot{\bm{x}} &= \bm{J}(\bm{\theta})\dot{\bm{\theta}} \\
    \bm{\tau} &= \bm{J}^{T}(\bm{\theta})\bm{F}
  \end{align}
  where $\bm{g}_{m}$ represents the mapping from $\bm{\theta}$ to $\bm{l}$, $\bm{g}_{j}$ represents the mapping from $\bm{\theta}$ to $\bm{x}$, $\bm{G}$ denotes the muscle Jacobian, and $\bm{J}$ denotes the joint Jacobian.
  Additionally, frequently used values include the minimum and maximum muscle tensions $\bm{f}^{\{min, max\}}$ and the minimum and maximum muscle velocities $\dot{\bm{l}}^{\{min, max\}}$.

  Next, we discuss joint angle estimation in musculoskeletal humanoids.
  As mentioned earlier, due to the complex joint structure of musculoskeletal humanoids, they often lack joint angle sensors.
  Therefore, it is necessary to estimate the joint angle from changes in muscle length.
  Here, we discuss a method that uses the Extended Kalman Filter (EKF) to estimate the joint angle from changes in muscle length \cite{ookubo2015learning}.
  The prediction and update equations of EKF are as follows,
  \begin{align}
    \nonumber\mbox{Predict:}\;\;\;\;\;\;\;\;\;\;\;\;& \\
    \Delta\bm{l}_{t} &= \bm{l}_{t}-\bm{l}_{t-1}& \\
    \bm{\theta}^{est}_{t|t-1} &= \bm{\theta}^{est}_{t-1|t-1} + \bm{G}^{+}(\bm{\theta}^{est}_{t-1|t-1})\Delta\bm{l}_t \\
    \bm{P}_{t|t-1} &= \bm{P}_{t-1|t-1} + \bm{Q}
  \end{align}
  \begin{align}
    \nonumber\mbox{Update:}\;\;\;\;\;\;\;\;\;\;\;\;& \\
    \bm{e}_{t} &= \bm{l}_{t} - \bm{g}_{m}(\bm{\theta}^{est}_{t|t-1}) \label{equation:observation-normal} \\
    \bm{G}_{t} &= \left.\frac{\partial \bm{g}_{m}}{\partial \bm{\theta}}\right|_{\bm{\theta} = \bm{\theta}^{est}_{t|t-1}} = \bm{G}(\bm{\theta}^{est}_{t|t-1}) \\
    \bm{S}_{t} &= \bm{G}_{t}\bm{P}_{t|t-1}\bm{G}^T_{t} + \bm{R} \\
    \bm{K}_{t} &= \bm{P}_{t|t-1}\bm{G}^T_{t}\bm{S}^{-1}_{t} \\
    \bm{\theta}^{est}_{t|t} &= \bm{\theta}^{est}_{t|t-1} + \bm{K}_{t}\bm{e}_{t} \\
    \bm{P}_{t|t} &= (\bm{I}-\bm{K}_{t}\bm{G}_{t})\bm{P}_{t|t-1}
  \end{align}
  where $t$ is the time step, $\bm{G}^{+}$ is the pseudoinverse of the muscle Jacobian, $\Delta\bm{l}$ is the change in the actual muscle length, $\bm{P}$ is the covariance matrix of the estimation error of the joint angle, $\bm{Q}$ is the covariance matrix of the error in the time transition, $\bm{e}$ is the observation residual, $\bm{R}$ is the covariance matrix in the observation space, $\bm{S}$ is the covariance matrix of the observation residual, and $\bm{K}$ is the Kalman gain.
  By repeating these predictions and updates, the obtained $\bm{\theta}^{est}_{t|t}$ becomes the current estimated joint angle $\bm{\theta}^{est}$.
  However, $\bm{\theta}^{est}$ obtained here is greatly affected by factors that cannot be modeled, such as muscle stretch and friction, resulting in an error from the actual joint angle.
  Therefore, the estimated value is corrected using the recognition of AR markers through vision to estimate a joint angle closer to the actual one \cite{kawaharazuka2018online}.
  For example, when measuring the joint angle of the arm, an AR marker is attached to the end of the hand, and its position is denoted as $\bm{p}_{marker}$.
  The previously estimated joint angle $\bm{\theta}^{est}$ is used as the initial value $\bm{\theta}^{init}$, and the inverse kinematics is solved as follows concerning the target coordinate $\bm{p}^{ref}=\bm{p}_{marker}$,
  \begin{align}
    \bm{\theta}^{est'} = \textrm{IK}(\bm{p}^{ref}=\bm{p}_{marker}, \bm{\theta}^{init}=\bm{\theta}^{est}) \label{eq:ik-vision}
  \end{align}
  where IK denotes the inverse kinematics, and $\bm{\theta}^{est'}$ represents the estimated joint angle corrected by vision.
  The $\bm{\theta}^{est'}$ obtained here can be used as the joint angle measured on the actual robot for learning and other purposes.

  Finally, the method to find the muscle tension $\bm{f}$ that can realize a given joint torque $\bm{\tau}$ is described.
  The muscle tension $\bm{f}$ that satisfies the desired joint torque $\bm{\tau}$ at $\bm{\theta}$ can be calculated by solving the following quadratic programming,
  \begin{align}
    \underset{\bm{f}}{\textrm{minimize}}&\;\;\;\;\;\;\;\;\;\;\;\;\;\;\;\;\;\bm{f}^{T}\bm{W}_{1}\bm{f}\label{eq:tension-calc-1}\\
    \textrm{subject to}&\;\;\;\;\;\;\;\;\;\;\;\; \bm{\tau} = -\bm{G}^{T}(\bm{\theta})\bm{f}\nonumber\\
    &\;\;\;\;\;\;\;\;\;\;\;\; \bm{f}^{min} \leq \bm{f} \leq \bm{f}^{max}\nonumber
  \end{align}
  where $\bm{W}_{1}$ is the weight matrix.
  However, there may be cases where a muscle tension $\bm{f}$ that perfectly realizes $\bm{\tau}$ does not exist.
  In such cases, the formulation is modified to allow for error as follows \cite{kawamura2016jointspace},
  \begin{align}
    \underset{\bm{f}}{\textrm{minimize}}&\;\;\;\bm{f}^{T}\bm{W}_{1}\bm{f} + (\bm{G}^{T}\bm{f}+\bm{\tau})^{T}W_{2}(\bm{G}^{T}\bm{f}+\bm{\tau})\label{eq:tension-calc-2}\\
    \textrm{subject to}&\;\;\;\;\;\;\;\;\;\;\;\;\;\;\;\;\;\;\;\;\;\;\;\;\bm{f}^{min} \leq \bm{f} \leq \bm{f}^{max}\nonumber
  \end{align}
  where $\bm{W}_{2}$ is the weight matrix.
  Note that the robots handled in this study basically operate under muscle length control, so this method is not used for muscle tension control.
}%
{%
  まず, 本研究で扱う筋骨格ヒューマノイドの基本構造について述べる.
  \figref{figure:structure}に示すように, 筋骨格構造は骨, 関節, 筋アクチュエータ, ワイヤ, 非線形弾性要素からなる.
  筋アクチュエータはモータの先に取り付けられたプーリによってワイヤを巻き取ることで関節を駆動する.
  ワイヤの先端には非線形弾性要素が取り付けられており, 筋肉の非線形弾性を模倣している.
  ワイヤはDyneema等の摩擦に強い化学繊維を用いる場合が多い.
  筋アクチュエータにはインクリメンタルエンコーダ, 張力測定ユニット, 温度センサが取り付けられており, それぞれから筋長$l$, 筋張力$f$, 筋温度$c$を計測することができる.
  必ず最初にロボットの関節角度を初期位置に戻し, そこでの筋長を0としてエンコーダの値を初期化する.
  ロボットによっては(例えばMusashi \cite{kawaharazuka2019musashi})関節にエンコーダが取り付けられている場合があり, 関節角度$\theta$が計測可能である.
  一方で, 人間の球関節や肩甲骨のような複雑な関節まで模倣する場合は(例えばKengoro \cite{asano2016kengoro}), 関節角度センサを入れることができない.
  その場合でも, 視覚センサによって手先の位置を計測することができれば, 筋長$l$の変化と手先の位置の関係から関節角度$\theta$を推定することができる\cite{kawaharazuka2018online}.
  筋肉は関節に対して冗長に配置されており, 一関節筋だけでなく, 複数の関節にまたがる筋肉も存在する.

  次に, 筋骨格ヒューマノイドにおける関節と筋肉の関係について基本的な数式を示す.
  筋数を$M$, 関節数を$N$, 筋張力を$\bm{f}$, 関節トルクを$\bm{\tau}$, 手先力を$\bm{F}$, 筋長を$\bm{l}$, 関節角度を$\bm{\theta}$, 手先位置を$\bm{x}$として, 以下の数式が一般的に成り立つ.
  \begin{align}
    \bm{l} &= \bm{g}_{m}(\bm{\theta}) \\
    \dot{\bm{l}} &= \bm{G}(\bm{\theta})\dot{\bm{\theta}} \\
    \bm{\tau} &= -\bm{G}^{T}(\bm{\theta})\bm{f} \\
    \bm{x} &= \bm{g}_{j}(\bm{\theta}) \\
    \dot{\bm{x}} &= \bm{J}(\bm{\theta})\dot{\bm{\theta}} \\
    \bm{\tau} &= \bm{J}^{T}(\bm{\theta})\bm{F}
  \end{align}
  ここで, $\bm{g}_{m}$は$\bm{\theta}$から$\bm{l}$への写像, $\bm{g}_{j}$は$\bm{\theta}$から$\bm{x}$への写像, $\bm{G}$は筋長ヤコビアン, $\bm{J}$は関節ヤコビアンを表す.
  また, よく使う値として, 筋張力$\bm{f}$の最小値と最大値$\bm{f}^{\{min, max\}}$, 筋速度$\dot{\bm{l}}$の最小値と最大値$\dot{\bm{l}}^{\{min, max\}}$がある.

  次に, 頻繁に登場する筋骨格ヒューマノイドの関節角度推定について述べる.
  前述のように, 筋骨格ヒューマノイドは複雑な関節構造から, 関節角度センサを持たない場合が多い.
  そのため, 筋長の変化から関節角度を推定する必要がある.
  ここでは筋長の変化から拡張カルマンフィルタを用いて関節角度を推定する手法\cite{ookubo2015learning}について述べる.
  拡張カルマンフィルタの予測式と更新式は以下である.
  \begin{align}
    \nonumber\mbox{Predict:}\;\;\;\;\;\;\;\;\;\;\;\;& \\
    \Delta\bm{l}_{t} &= \bm{l}_{t}-\bm{l}_{t-1}& \\
    \bm{\theta}^{est}_{t|t-1} &= \bm{\theta}^{est}_{t-1|t-1} + \bm{G}^{+}(\bm{\theta}^{est}_{t-1|t-1})\Delta\bm{l}_t \\
    \bm{P}_{t|t-1} &= \bm{P}_{t-1|t-1} + \bm{Q} \\
    \nonumber\mbox{Update:}\;\;\;\;\;\;\;\;\;\;\;\;& \\
    \bm{e}_{t} &= \bm{l}_{t} - \bm{g}_{m}(\bm{\theta}^{est}_{t|t-1}) \label{equation:observation-normal} \\
    \bm{G}_{t} &= \left.\frac{\partial \bm{g}_{m}}{\partial \bm{\theta}}\right|_{\bm{\theta} = \bm{\theta}^{est}_{t|t-1}} = \bm{G}(\bm{\theta}^{est}_{t|t-1}) \\
    \bm{S}_{t} &= \bm{G}_{t}\bm{P}_{t|t-1}\bm{G}^T_{t} + \bm{R} \\
    \bm{K}_{t} &= \bm{P}_{t|t-1}\bm{G}^T_{t}\bm{S}^{-1}_{t} \\
    \bm{\theta}^{est}_{t|t} &= \bm{\theta}^{est}_{t|t-1} + \bm{K}_{t}\bm{e}_{t} \\
    \bm{P}_{t|t} &= (\bm{I}-\bm{K}_{t}\bm{G}_{t})\bm{P}_{t|t-1}
  \end{align}
  ここで, $t$がタイムステップ数, $\bm{G}^{+}$が筋長ヤコビアンの擬似逆行列, $\Delta\bm{l}$が実機の筋長変化分, $\bm{P}$が関節角度推定値の誤差の共分散行列, $\bm{Q}$が時間遷移に関する誤差の共分散行列, $\bm{e}$が観測残差, $\bm{R}$が観測空間の共分散行列, $\bm{S}$が観測残差の共分散行列, $\bm{K}$がカルマンゲインである.
  これらの予測と更新を繰り返すことで, 得られた$\bm{\theta}^{est}_{t|t}$が現在の関節角度推定値$\bm{\theta}^{est}$となる.
  一方, ここで得られた$\bm{\theta}^{est}$は筋の伸びや摩擦等, モデル化できていない要素の影響を大きく受け, 実機関節角度とは誤差がある.
  そのため, 視覚を使ったARマーカの認識によりこの推定値を補正し, より実機に近い関節角度を推定する\cite{kawaharazuka2018online}.
  例えば腕の関節角度を測定したい場合, その手の先端にARマーカ等を取り付け, その位置を$\bm{p}_{marker}$とする.
  先程の関節角度推定値$\bm{\theta}^{est}$を初期値$\bm{\theta}^{init}$として, 指令座標$\bm{p}^{ref}=\bm{p}_{marker}$に対して以下のように逆運動学を解く.
  \begin{align}
    \bm{\theta}^{est'} = \textrm{IK}(\bm{p}^{ref}=\bm{p}_{marker}, \bm{\theta}^{init}=\bm{\theta}^{est}) \label{eq:ik-vision}
  \end{align}
  ここで, IKは逆運動学を, $\bm{\theta}^{est'}$は視覚により補正された実機関節角度の推定値を表す.
  ここで得られた$\bm{\theta}^{est'}$を, 実機で測定された関節角度$\bm{\theta}$として学習等に用いることが可能である.

  最後に, ある関節トルク$\bm{\tau}$を実現可能な筋張力$\bm{f}$を求める方法について述べる.
  $\bm{\theta}$において実現したい関節トルク$\bm{\tau}$を満たす筋張力$\bm{f}$を, 以下のような二次計画法を解くことで計算する.
  \begin{align}
    \underset{\bm{f}}{\textrm{minimize}}&\;\;\;\;\;\;\;\;\;\;\;\;\;\;\;\;\;\bm{f}^{T}\bm{W}_{1}\bm{f}\label{eq:tension-calc-1}\\
    \textrm{subject to}&\;\;\;\;\;\;\;\;\;\;\;\; \bm{\tau} = -\bm{G}^{T}(\bm{\theta})\bm{f}\nonumber\\
    &\;\;\;\;\;\;\;\;\;\;\;\; \bm{f}^{min} \leq \bm{f} \leq \bm{f}^{max}\nonumber
  \end{align}
  ここで, $\bm{W}_{1}$は重み行列を表す.
  一方で, 完璧に$\bm{\tau}$を実現する$\bm{f}$が存在しない場合があり, その場合は以下のように定式化を変更し誤差を許容する\cite{kawamura2016jointspace}.
  \begin{align}                                 
    \underset{\bm{f}}{\textrm{minimize}}&\;\;\;\bm{f}^{T}\bm{W}_{1}\bm{f} + (\bm{G}^{T}\bm{f}+\bm{\tau})^{T}W_{2}(\bm{G}^{T}\bm{f}+\bm{\tau})\label{eq:tension-calc-2}\\
    \textrm{subject to}&\;\;\;\;\;\;\;\;\;\;\;\;\;\;\;\;\;\;\;\;\;\;\;\;\bm{f}^{min} \leq \bm{f} \leq \bm{f}^{max}\nonumber
  \end{align}
  ここで, $\bm{W}_{2}$は重み行列を表す.
  なお, 本研究で扱うロボットは基本的に筋長制御で動くため, この方法を筋張力制御に用いているわけではない点には注意していただきたい.
}%

\subsection{Characteristics of Muscles in Musculoskeletal Humanoids}\label{subsec:characteristics}
\switchlanguage%
{%
  In this study, the characteristics of muscles in musculoskeletal humanoids were classified into the following five categories.
  \begin{itemize}
    \item Redundancy
    \item Independency
    \item Anisotropy
    \item Variable Moment Arm
    \item Nonlinear Elasticity
  \end{itemize}
  Redundancy is the most important characteristic of muscles, indicating that multiple muscles can act on a single joint.
  Independency refers to the ability to arrange joints and muscles independently, unlike the axis-driven robots where actuators and joints are coupled.
  Anisotropy describes the nature of muscles that can exert force only in the contraction direction and can loosen in the opposite direction.
  Variable Moment Arm indicates that, due to its mimicry of the human body, the moment arm of the muscles to the joint varies depending on the joint angle.
  Nonlinear Elasticity refers to the presence of nonlinear elastic elements attached to the muscles, which is unique to the musculoskeletal structure.

  Note that Redundancy refers to both the property where muscles act antagonistically on a single joint and the property where multiple agonist and antagonist muscles can coexist for the same joint.
}%
{%
  本研究では, 筋骨格ヒューマノイドにおける筋肉の特性を以下の5つに分類した.
  \begin{itemize}
    \item Redundancy
    \item Independency
    \item Anisotropy
    \item Variable Moment Arm
    \item Nonlinear Elasticity
  \end{itemize}
  ``Redundancy''とは, 筋肉の最も重要な特性であり, 1つの関節に対して複数の筋肉が作用できる様子を表す.
  ``Independency''とは, 軸駆動型ロボットのアクチュエータと関節がカップリングしている様子とは異なり, 関節と筋肉が独立に配置可能な様を表す.
  ``Anisotropy''とは, 筋肉が収縮方向しか力を発揮することができず, 逆に反対方向に緩むことが可能な様を表す.
  ``Variable Moment Arm''とは, 人体模倣ゆえに, 関節に対する筋のモーメントアームが関節角度によって変化する様を表す.
  ``Nonlinear Elasticity''とは, これら筋骨格構造に特有な, 筋肉に取り付けた非線形弾性要素の存在を表す.

  なお, ここでいうRedundancyは, 1つの関節に対して拮抗に筋肉が作用するという性質と, 主動筋・拮抗筋もそれぞれ複数存在することができるという性質の両方を含む.
}%

\begin{figure}[t]
  \centering
  \includegraphics[width=0.95\columnwidth]{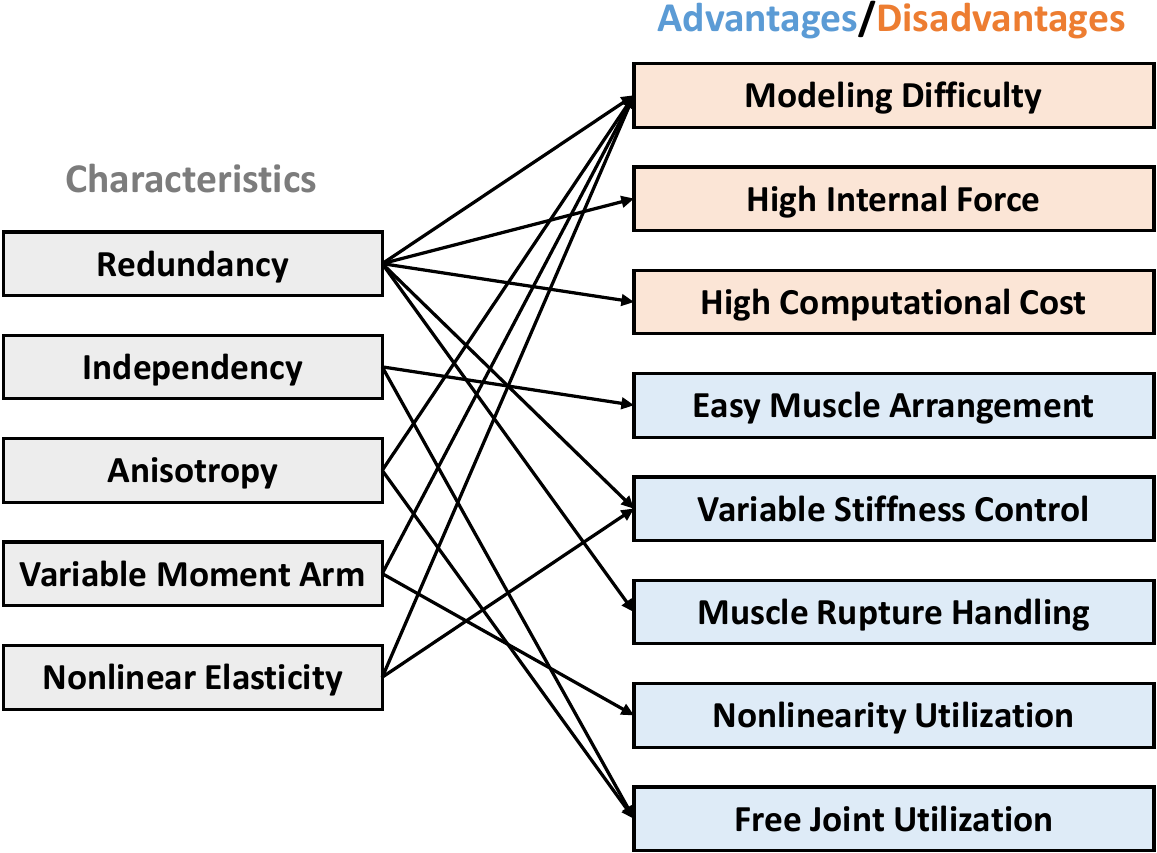}
  \caption{The relationship between the characteristics of muscles and the advantages/disadvantages of the musculoskeletal structure.}
  \label{figure:relationship}
  \vspace{-1.0ex}
\end{figure}

\begin{table*}[t]
  \centering
  \caption{The relationship between the control research on Kengoro/Musashi and advantages/disadvantages of the musculoskeletal structure.}
  \scalebox{0.85}{
    \begin{tabular}{|c||c|c|c|c|c|c|c|c|c|} \hline
               & Modeling   & High           & High               & Easy Muscle & Variable          & Muscle Rupture & Nonlinearity & Free Joint \\
      Research & Difficulty & Internal Force & Computational Cost & Arrangement & Stiffness Control & Handling       & Utilization   & Utilization \\ \hline\hline
      Antagonist Inhibition \cite{kawaharazuka2017antagonist} & \checkmark & \checkmark &            &            &            &            &            &            \\ \hline
      Thermal Control \cite{kawaharazuka2020thermo}           &            & \checkmark &            &            &            &            &            &            \\ \hline
      Relaxation Control \cite{kawaharazuka2019relax}         &            & \checkmark &            &            &            &            &            &            \\ \hline
      Stretch Reflex \cite{kawaharazuka2020stretch}           & \checkmark &            &            &            &            &            &            &            \\ \hline
      Joint-muscle Mapping \cite{kawaharazuka2020autoencoder} & \checkmark & \checkmark &            &            &            &            &            &            \\ \hline
      Predictive Model \cite{kawaharazuka2023dpmpb}           & \checkmark &            &            &            &            &            &            &            \\ \hline
      Muscle Grouping \cite{kawaharazuka2021grouping}         &            &            & \checkmark &            &            &            &            &            \\ \hline
      Muscle Addition \cite{kawaharazuka2022additional}       &            &            &            & \checkmark &            &            &            &            \\ \hline
      Variable Stiffness \cite{kawaharazuka2019longtime}      &            &            &            &            & \checkmark &            &            &            \\ \hline
      Muscle Rupture \cite{kawaharazuka2022redundancy}        &            &            &            &            &            & \checkmark &            &            \\ \hline
      Design Optimization \cite{kawaharazuka2021redundancy}   &            &            &            & \checkmark &            & \checkmark &            &            \\ \hline
      Nonlinear Estimation \cite{kawaharazuka2018estimator}   &            &            &            &            &            &            & \checkmark &            \\ \hline
      Maximum Speed \cite{kawaharazuka2020speed}              &            &            &            &            &            &            &            & \checkmark \\ \hline
    \end{tabular}
    \label{table:relationship}
    }
\end{table*}

\subsection{Advantages/Disadvantages of the Musculoskeletal Structure} \label{subsec:advantage}
\switchlanguage%
{%
  The five muscle characteristics described in \secref{subsec:characteristics} mainly result in the following eight advantages and disadvantages.
  \begin{itemize}
    \item Modeling Difficulty
    \item High Internal Force
    \item High Computational Cost
    \item Easy Muscle Arrangement
    \item Variable Stiffness Control
    \item Muscle Rupture Handling
    \item Nonlinearity Utilization
    \item Free Joint Utilization
  \end{itemize}
  The top three are disadvantages, and the bottom five are advantages.
  The relationships between muscle characteristics and their advantages and disadvantages are shown in \figref{figure:relationship}.

  First, Modeling Difficulty arises from various muscle characteristics such as Redundancy, Anisotropy, Variable Moment Arm, and Nonlinear Elasticity.
  While axis-driven robots and wire-driven systems with constant moment arms are relatively easy to model, the nonlinear, anisotropic, and redundant nature of muscles makes modeling more challenging.
  On the other hand, humans are able to skillfully move their complex bodies, so the understanding of this body structure and its control principles is an important research topic.
  Next, High Internal Force is caused by Redundancy.
  When muscles are redundantly arranged around a joint, agonist and antagonist muscles can be created, potentially generating high internal forces when they oppose each other.
  High Computational Cost also results from Redundancy.
  Due to redundancy, the relationship between joints and muscles becomes complex.
  For example, calculating the muscle Jacobian or estimating the relationship between joint angle and muscle length requires significant computational resources.
  Easy Muscle Arrangement is derived from Independency.
  The ability to arrange muscles independently of the joints allows for the free adjustment of muscle placement and the moment arm.
  It is not difficult to add new muscles as needed though this is a feature not present in humans.
  Variable Stiffness Control arises from Redundancy and Nonlinear Elasticity.
  This is one of the most important characteristics of musculoskeletal structures, allowing for the immediate adaptation to various environments by changing body stiffness at the hardware level rather than through software.
  Muscle Rupture Handling is also a result of Redundancy.
  With muscles redundantly arranged around the joints, if one muscle is damaged, other muscles can compensate for its function, allowing the task to continue.
  While studies have shown that robustness to muscle rupture is not very high in humans \cite{kutch2011dysfunction}, it is possible to freely design muscle configurations in musculoskeletal humanoids to ensure robustness against muscle rupture.
  Nonlinearity Utilization results from Variable Moment Arm.
  By leveraging the variation in the moment arm of muscles according to the joint angle, nonlinear changes in muscle length can be utilized for control and state estimation.
  When referring to Nonlinearity, one might think of nonlinear elastic elements; however, here it specifically refers to the nonlinearity of the muscle moment arms.
  Finally, Free Joint Utilization arises from Independency and Anisotropy.
  By relaxing the muscles to free the joint, it is possible to move the joint flexibly, ignoring the backdrivability of the actuator.
}%
{%
  \secref{subsec:characteristics}で述べた5つの筋肉の特徴は, 主に以下の8の利点・欠点を生み出す.
  \begin{itemize}
    \item Modeling Difficulty
    \item High Internal Force
    \item High Computational Cost
    \item Easy Muscle Arrangement
    \item Variable Stiffness Control
    \item Muscle Rupture Handling
    \item Nonlinearity Utilization
    \item Free Joint Utilization
  \end{itemize}
  なお, 上3つが欠点, 下5つが利点である.
  これら, 筋肉の特性と利点・欠点の関係を\figref{figure:relationship}に示す.

  まず, ``Modeling Difficulty''はRedundancy, Anisotropy, Variable Moment Arm, Nonlinear Elasticityという様々な筋肉の特性によって生じる.
  軸駆動型のロボットやモーメントアーム一定のワイヤ駆動系であれば非常にモデル化が簡単である一方で, 非線形で異方性や冗長性を持つ筋肉のモデル化は難しい.
  同時に, 人間はそのような複雑な体でも巧みに身体を動かしており, その身体構造と制御原理の理解は重要な課題である.
  次に, ``High Internal Force''はRedundancyによって生じる.
  関節に対して冗長な筋肉が配置されることで, 主動筋と拮抗筋が生まれ, 場合によってはこれらが引き合うことで高い内力が発生する危険がある.
  次に, ``High Computational Cost''はRedundancyによって生じる.
  冗長性により, 関節と筋の間の関係が複雑になり, 例えば筋長ヤコビアンを計算したり, 関節角度と筋長の関係を推定したりするためには, 多くの計算リソースが必要となる.
  次に, ``Easy Muscle Arrangement''はIndependencyによって生じる.
  筋肉を関節と独立に配置可能な性質によって, 筋肉の配置を自由に変更し, モーメントアームを調整することが可能である.
  人間にはない機能であるが, 必要に応じて新しい筋肉を追加することも難しくない.
  次に, ``Variable Stiffness Control''はRedundancyとNonlinear Elasticityによって生じる.
  これは筋骨格構造を語る上で最も重要な特性のうちの一つであり, ソフトウェアではなくハードウェアとして身体剛性を変化させることで, 様々な環境に即座に対応することが可能である.
  次に, ``Muscle Rupture Handling''はRedundancyによって生じる.
  筋肉が関節に対して冗長に配置されることで, 1つの筋肉が破損しても他の筋肉がその機能を補い, 継続的にタスクを実行することができる.
  人間において筋破断に対するrobustnessはあまり高くないという研究もあるが\cite{kutch2011dysfunction}, 筋骨格ヒューマノイドにおいては筋破断に頑健なワイヤ配置を自由に設計することが可能である.
  次に, ``Nonlinearity Utilization''はVariable Moment Armによって生じる.
  関節角度に応じて筋のモーメントアームが変化することを利用して, 非線形な筋長変化を制御や状態推定に利用することができる.
  Nonlinearityというと非線形弾性要素が思い浮かぶかもしれないが, ここでは筋のモーメントアームの非線形性を指している.
  最後に, ``Free Joint Utilization''はIndependencyとAnisotropyによって生じる.
  筋を伸ばして関節をフリーにすることで, アクチュエータのバックドライバビリティを無視して, 自由に柔らかく関節を動かすことが可能である.
}%

\begin{figure*}[t]
  \centering
  \includegraphics[width=1.95\columnwidth]{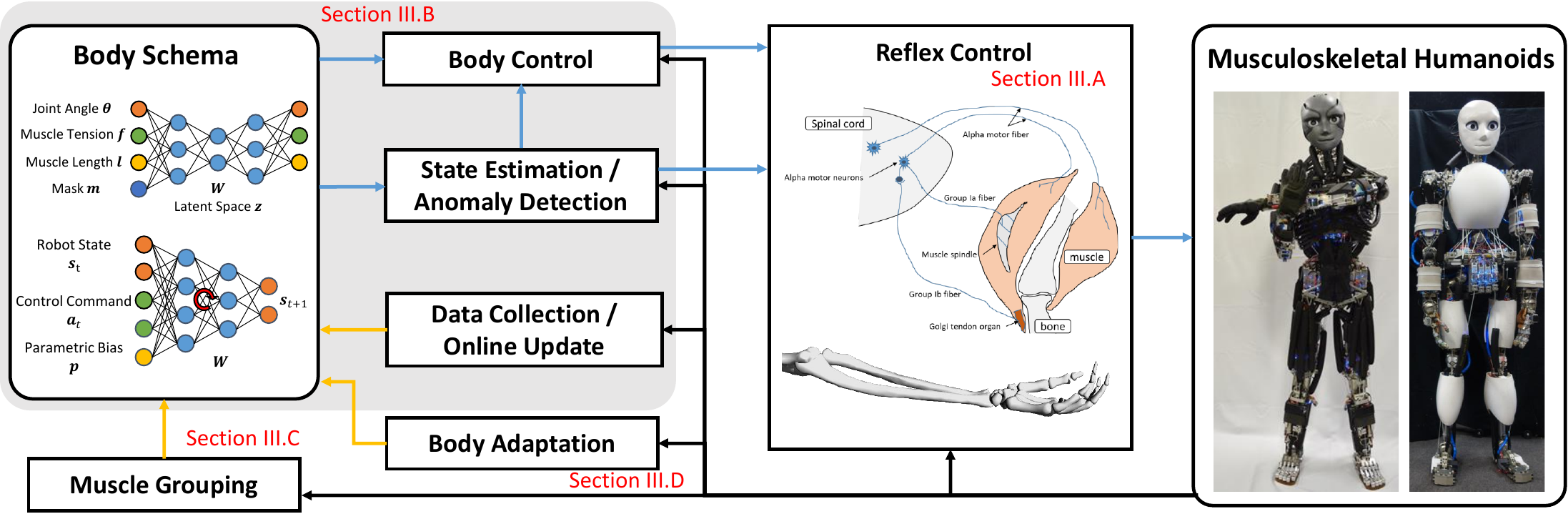}
  \caption{The overview of the system which manages and utilizes advantages/disadvantages of the musculoskeletal structure. The components include reflex control (\secref{subsec:reflex}), body schema learning (\secref{subsec:learning}), muscle grouping for body schema (\secref{subsec:grouping}), and body schema adaptation (\secref{subsec:adaptation}).}
  \label{figure:system}
  \vspace{-1.0ex}
\end{figure*}

\subsection{Management and Utilization of the Musculoskeletal Structure on Kengoro and Musashi} \label{subsec:empirical}
\switchlanguage%
{%
  As examples of research conducted with the musculoskeletal humanoids Kengoro and Musashi, we illustrate how to manage these various disadvantages and how to utilize the advantages.

  First, let's briefly describe the body structures of Kengoro and Musashi.
  As shown in \figref{figure:robots}, Kengoro and Musashi are robots with human-like musculoskeletal structures.
  Kengoro currently has 124 muscles, while Musashi has 74 muscles in its body.
  Kengoro was developed to closely mimic human anatomy, featuring a structure very similar to the human skeleton, including the spine, ribs, scapula, radius, and ulna.
  Therefore, joint angle sensors are not installed, and as described in \secref{subsec:structure}, it is necessary to estimate the current joint angle from muscle length changes and vision sensors.
  On the other hand, Musashi is a robot designed for ease of design modification by modularizing joints and muscles without aiming for a detailed human mimicry.
  Encoders are installed within its joint modules, allowing direct measurement of joint angle, making it a platform for research on learning control.
  In addition to numerous contact sensors on the hands and feet, Musashi is equipped with movable eyeballs, a high-precision camera, and an auditory system, enabling control by integrating various sensor information.

  A table that correlates the control research conducted on Kengoro and Musashi with the advantages and disadvantages of muscles is shown in \tabref{table:relationship}.
  For Modeling Difficulty, low-level reflex controls (antagonist inhibition control \cite{kawaharazuka2017antagonist} and stretch reflex control \cite{kawaharazuka2020stretch}) and high-level learning controls (learning of joint-muscle mapping \cite{kawaharazuka2020autoencoder} and deep predictive model \cite{kawaharazuka2023dpmpb}) can be effectively employed.
  Note that the low-level reflex controller is a controller that accesses a limited number of sensor values and responds quickly at a high frequency (greater than 100 Hz).
  In contrast, the high-level learning controller is a controller that accesses a wider range of modalities and operates at a slower frequency (around 10 Hz is sufficient).
  High Internal Force can also be managed by combining low-level reflex controls (antagonist inhibition control \cite{kawaharazuka2017antagonist}, muscle thermal control \cite{kawaharazuka2020thermo}, and muscle relaxation control \cite{kawaharazuka2019relax}) and high-level learning control (learning of joint-muscle mapping \cite{kawaharazuka2020autoencoder}).
  These reflex controls can address modeling errors and internal forces in the short term, while learning control can correct modeling errors and internal forces in the long term.
  To address High Computational Cost, the computational load of joint-muscle mapping can be reduced by appropriately grouping muscles and joints \cite{kawaharazuka2021grouping}.
  Easy Muscle Arrangement allows for the addition of muscles depending on the task, requiring the body schema to be relearned depending on the body change \cite{kawaharazuka2022additional}.
  It is also possible to design muscle placements according to the task \cite{kawaharazuka2021redundancy}.
  Variable Stiffness Control enables the robot to perform tasks while changing body stiffness according to various environments \cite{kawaharazuka2019longtime}.
  Muscle Rupture Handling makes it possible to respond to muscle rupture, allowing for continuous task execution while sequentially updating the body schema \cite{kawaharazuka2022redundancy}.
  It is also possible to select muscle arrangements that are easier to handle in case of muscle rupture \cite{kawaharazuka2021redundancy}.
  Nonlinearity Utilization allows for the use of nonlinear muscle length changes corresponding to joint angle changes, specifically the nonlinear changes in muscle Jacobian, enabling absolute joint angle estimation using only relative muscle length changes \cite{kawaharazuka2018estimator}.
  Normally, relative changes in muscle length only provide information about changes in joint angle, requiring initialization by setting the muscle length to zero relative to the initial joint angle.
  However, with this approach, such initialization is no longer necessary.
  Free Joint Utilization reduces the influence of backdrivability by freeing the joints, allowing for rapid movements that ignore the maximum speed of actuators \cite{kawaharazuka2020speed}.

  In the following sections, we organize these studies in a more generalized form and present the integrated overall system.
  Please refer to each study for details, as we will summarize the outline of each component to maintain coherence.
}%
{%
  これまで我々が開発してきた筋骨格ヒューマノイドKengoroとMusashiで行われてきた研究を例として, この多様な欠点をどう管理し, 利点をどう利用することが可能なのか, その概要を示す.

  まず, KengoroとMusashiの身体構造について簡単に述べる.
  \figref{figure:robots}に示すように, KengoroとMusashiは人間同様な筋骨格構造を持つロボットである.
  Kengoroは現在124本の筋, Musashiは74本の筋を体内に持つ.
  Kengoroは詳細な人体模倣を目指し開発されたロボットであり, 背骨や肋骨, 肩甲骨, 橈骨尺骨など, 人間の骨格に非常に近い構造を持つ.
  そのため, 関節角度センサが取り付けられておらず, \secref{subsec:structure}で述べたように, 筋長変化と視覚から現在の関節角度を推定する必要がある.
  一方でMusashiは, 詳細な人体模倣までは目指さず, 関節や筋をモジュール化することで容易な設計変更を可能としたロボットである.
  関節モジュール内にはエンコーダが取り付けられており, 関節角度を直接計測することが可能で, 学習制御研究のプラットフォームとして利用されてきた.
  手先や足先に多数の接触センサが取り付けられているだけでなく, 可動する眼球と高精度カメラ, 聴覚システムまで搭載されており, 様々なセンサ情報を統合した制御が可能である.

  これらKengoroとMusashiで行われてきた制御研究と筋肉の利点・欠点を対応させた表を\tabref{table:relationship}に示す.
  ``Modeling Difficulty''に対しては, 低レイヤの高速な反射制御(拮抗筋抑制制御\cite{kawaharazuka2017antagonist}や伸長反射制御\cite{kawaharazuka2020stretch})と高レイヤの学習制御(関節-筋空間マッピング\cite{kawaharazuka2020autoencoder}, 深層予測モデル制御\cite{kawaharazuka2023dpmpb})が効果的に対応可能である.
  なお, 低レイヤ反射制御器は少数のセンサ値にアクセスし速い周期で(> 100 Hz)で素早く反応する制御器であり, 高レイヤ学習制御器はより多くのモダリティにアクセスし遅い周期で(10 Hz程度で十分)で実行される制御器を表しています.
  ``High Internal Force''に対しても同様に, 反射型制御(拮抗筋抑制制御\cite{kawaharazuka2017antagonist}, 筋温度制御\cite{kawaharazuka2020thermo}, 筋弛緩制御\cite{kawaharazuka2019relax}, 伸長反射制御\cite{kawaharazuka2020stretch})と学習制御(関節-筋空間マッピング\cite{kawaharazuka2020autoencoder})を複合することで対処することができる.
  これら反射制御は短期的にモデル化誤差や内力に対処することが可能であり, 学習制御は長期的にモデル化誤差や内力を修正していくことができる.
  ``High Computational Cost''に対しては, 筋と関節を適切にグルーピングすることで, 関節-筋空間マッピングの計算量を削減することができる\cite{kawaharazuka2021grouping}.
  ``Easy Muscle Arrangement''により, タスクに応じて筋を追加することができ, それに依存して身体図式を学習し直す必要がある\cite{kawaharazuka2022additional}.
  また, タスクに応じて筋配置を変化させることも可能である\cite{kawaharazuka2021redundancy}.
  ``Variable Stiffness Control''により, 様々な環境に応じて身体剛性を変えながらタスクを行うことができる\cite{kawaharazuka2019longtime}.
  ``Muscle Rupture Handling''により, 筋の破断に対応することが可能であり, 身体図式を逐次的に更新しながら継続的にタスクを実行することが可能である\cite{kawaharazuka2022redundancy}.
  また, 筋破断に対応しやすい筋配置を選択することも可能である\cite{kawaharazuka2021redundancy}.
  ``Nonlinearity Utilization''により, 関節角度変化に対する非線形な筋長変化, つまり筋長ヤコビアンの非線形変化を利用し, 相対筋長変化のみを用いた絶対関節角度推定が可能である\cite{kawaharazuka2018estimator}.
  通常筋肉の相対変化のみでは関節角度変化しか分からないため初期関節角度に対する筋長を0として初期化する必要があるが, この操作が必要なくなる.
  ``Free Joint Utilization''により, 関節をフリーにすることでバックドライバビリティの影響を減らし, アクチュエータの最大速度を無視した素早い動作が可能である\cite{kawaharazuka2020speed}.

  以降では, これらの研究をより一般化した形で整理し, それらを統合した全体システムを示す.
  なお, 統一性を重視して各コンポーネントの概要をまとめるため, 詳細についてはそれぞれの研究を参照されたい.
}%

\section{Empirical Study on Kengoro and Musashi} \label{sec:detail}
\switchlanguage%
{%
  \figref{figure:system} shows an overview of the software system for musculoskeletal humanoids, summarizing various research.
  This system mainly comprises low-level reflex control and high-level body schema learning.
  Surrounding the body schema are components for data collection and its training, state estimation, anomaly detection, and body control using the body schema.
  At the same time, there are components for muscle grouping for the body schema aimed at reducing computational costs and body schema adaptation that incrementally responds to bodily changes.

  \secref{subsec:reflex} discusses low-level reflex control, \secref{subsec:learning} covers body schema learning and control, \secref{subsec:grouping} describes muscle grouping in the body schema, and \secref{subsec:adaptation} explains body schema adaptation.
}%
{%
  これまでの研究をまとめた, 筋骨格ヒューマノイドにおけるソフトウェアシステムの概要を\figref{figure:system}に示す.
  このシステムには主に低レイヤの反射制御, 高レイヤの身体図式学習が備わっている.
  身体図式を取り囲むように, データ収集と身体図式学習, 身体図式を用いた状態推定や異常検知, 身体制御のコンポーネントが存在する.
  それと同時に, 計算コストの低下を目指した身体図式における筋のグルーピング, 身体変化に逐次的に対応する身体図式適応のコンポーネントが存在する.
  本章では, \secref{subsec:reflex}で低レイヤの反射制御について, \secref{subsec:learning}で身体図式学習と制御について, \secref{subsec:grouping}で身体図式のグルーピングについて, \secref{subsec:adaptation}で身体図式の適応について述べる.
}%

\subsection{Reflex Control for Musculoskeletal Humanoids} \label{subsec:reflex}
\begin{figure}[t]
  \centering
  \includegraphics[width=0.95\columnwidth]{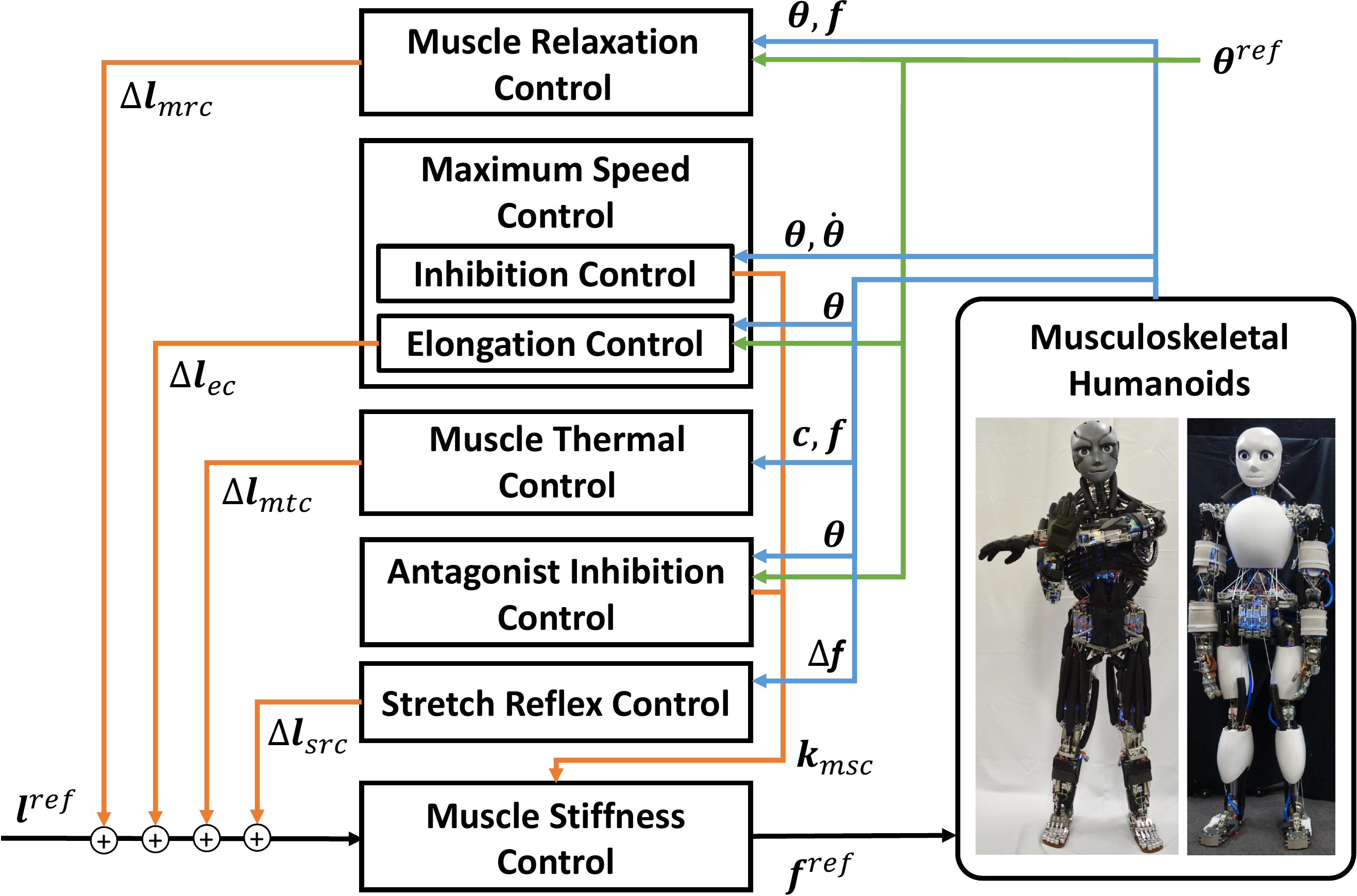}
  \caption{The overview of the reflex control system. Muscle relaxation control, maximum speed control, muscle thermal control, antagonist inhibition control, and stretch reflex control alter the muscle length and the muscle stiffness in muscle stiffness control.}
  \label{figure:reflex}
  \vspace{-1.0ex}
\end{figure}

\switchlanguage%
{%
  In musculoskeletal humanoids, it is difficult to control all the muscles from the upper layer.
  Therefore, controlling the muscle length and stiffness through low-level reflex control can enhance the robot's adaptability.
  Since musculoskeletal humanoids closely mimic the human body, it is also possible to directly implement reflex controls similar to those in humans.
  \figref{figure:reflex} shows an integrated reflex control system that includes stretch reflex control, antagonist inhibition control, muscle thermal control, maximum speed control, and muscle relaxation control.

  First, the lowest-level control of the musculoskeletal humanoids Kengoro and Musashi is called Muscle Stiffness Control (MSC) and can be expressed for each muscle $i$ as follows \cite{shirai2011stiffness},
  \begin{align}
    f^{ref}_{i} = f^{bias}_{i} + \textrm{max}(0, k_{msc, i}(l_{i}-l^{ref}_{i})) \label{eq:msc}
  \end{align}
  where $\bm{\cdot}_{i}$ represents the value of the $i$-th muscle, $\bm{f}^{ref}$ is the target muscle tension, $\bm{f}^{bias}$ is the bias term of muscle stiffness control, $\bm{k}_{msc}$ is the muscle stiffness coefficient, and $\bm{l}^{ref}$ is the target muscle length.
  In the following various reflex controls, the robot's behavior is modified by changing $\bm{k}_{msc}$ and $\bm{l}^{ref}$ in this muscle stiffness control.

  Let's briefly discuss human reflexes.
  In humans, muscle spindles are spindle-shaped organs attached in parallel to muscle fibers and serve as receptors that detect muscle length $l$ and velocity $\dot{l}$.
  Group Ia fibers from the muscle spindles are excitatory to the spinal cord's $\alpha$ motor neurons.
  When the muscle is suddenly stretched, the frequency of impulses from the muscle spindle increases, exciting the $\alpha$ motor neurons and causing the stretched muscle to contract.
  This forms a reflex loop known as the stretch reflex, a negative feedback system with muscle length as its output.
  In contrast, tendon organs are located at the ends of muscles.
  Unlike muscle spindles, which are arranged in parallel to muscle fibers, tendon organs are connected in series with muscle fibers.
  Group Ib fibers from tendon organs connect to the spinal cord's $\alpha$ motor neurons via inhibitory interneurons.
  When the muscle contracts and muscle tension increases, the frequency of impulses from the tendon organs rises, inhibiting the excitatory action on the $\alpha$ motor neurons.
  This forms a reflex loop known as the tendon reflex system, a negative feedback system with muscle tension as its output.
  Additionally, the Ia afferent fibers from the muscle spindles not only excite the motor neurons of the agonist muscle but also inhibit the motor neurons of the antagonist muscle via inhibitory neurons.
  This interaction between motor neurons of antagonist and agonist muscles is called reciprocal innervation.
  The stretch reflex control discussed below mimics the human stretch reflex, and the antagonist inhibition control mimics the human reciprocal innervation.
}%
{%
  筋骨格ヒューマノイドにおいて, 多数の筋を全て上位レイヤから制御することは困難であるため, 低レイヤの反射制御によって筋肉の長さや硬さを制御することで, ロボットの適応性を高めることができる.
  また, 筋骨格ヒューマノイドは人体を詳細に模倣しているがゆえに, 人間と同様の反射制御を直接実装することができる.
  これまで開発されてきた伸長反射制御\cite{kawaharazuka2020stretch}, 拮抗筋抑制制御\cite{kawaharazuka2017antagonist}, 筋温度制御\cite{kawaharazuka2020thermo}, 最大速度制御\cite{kawaharazuka2020speed}, 筋弛緩制御\cite{kawaharazuka2019relax}を統合した反射制御システムを\figref{figure:reflex}に示す.

  まず, 筋骨格ヒューマノイドKengoroとMusashiの最下層の制御はMuscle Stiffness Control (MSC)と呼ばれ, 各筋$i$に対して以下のように表現できる\cite{shirai2011stiffness}.
  \begin{align}
    f^{ref}_{i} = f^{bias}_{i} + \textrm{max}(0, k_{msc, i}(l_{i}-l^{ref}_{i})) \label{eq:msc}
  \end{align}
  ここで, $\bm{\cdot}_{i}$は$i$番目の筋の値, $\bm{f}^{ref}$は指令筋張力, $\bm{f}^{bias}$は筋剛性制御のバイアス項, $\bm{k}_{msc}$は筋剛性制御の筋剛性係数, $\bm{l}^{ref}$は指令筋長である.
  以下の多様な反射制御では, この筋剛性制御の$\bm{k}_{msc}$と$\bm{l}^{ref}$を変化させることでロボットの挙動を変化させる.

  ここで簡単に人間の反射について触れておく.
  人間において, 筋紡錘は筋繊維に平行に付着する紡錘形の器官であり, 筋の長さ$l$や短縮速度$\dot{l}$を検知する受容器である.
  筋紡錘からはIa群繊維が脊髄の$\alpha$運動ニューロンに直接興奮性接続されており, 急に筋が伸ばされると筋紡錘からのインパルス頻度が増大し, $\alpha$運動ニューロンを興奮させ伸ばされた筋が収縮する.
  これは伸長反射と呼ばれる反射ループを成し, 筋長を出力とする負のフィードバック系を成す.
  また, 腱器官は筋の端に配置されており, 筋紡錘が筋繊維と平行に配置されているのに対し, 筋繊維と直列に接続する。
  腱器官からはIb群繊維が抑制性介在ニューロンを経て脊髄の$\alpha$運動ニューロンに接続しているため, 筋が収縮し筋張力が増すと腱器官からのインパルス頻度が増大し, $\alpha$運動ニューロンの興奮作用が抑制される.
  これは腱反射系とよばれる反射ループを成し, 筋張力を出力とする負のフィードバック系を成す.
  そして, 筋紡錘からのIa求心性繊維は主動筋の運動ニューロンを興奮させると同時に, 抑制性のニューロンを介して拮抗筋の運動ニューロンを抑制する神経回路が用意されている.
  このような拮抗筋と主動筋の運動ニューロン間の相互作用を相反性神経支配という.
  以降での伸長反射制御は人間の伸長反射を, 拮抗筋抑制制御は人間の相反生神経支配を模倣した制御である.
}%

\switchlanguage%
{%
  \subsubsection{Stretch Reflex Control (SRC)}
  Stretch Reflex Control (SRC) is a control that mimics the human stretch reflex, as mentioned earlier.
  It contracts each muscle $i$ when its muscle length $l_i$ is suddenly stretched.
  In other words, the stretch reflex occurs when the change in muscle length $\Delta{l}_i = l_{i, t+1} - l_{i, t}$ ($l_{i, t}$ represents the muscle length at time $t$) exceeds a threshold $C_{src}$ (where $C_{src}$ is a constant).
  However, as described in \secref{subsec:structure}, musculoskeletal humanoids often include nonlinear elastic elements.
  In this case, even if the muscle is suddenly stretched due to an impact, the deformation of the nonlinear elastic elements, which are more flexible than the motor and can absorb the shock, becomes dominant.
  The relationship between the muscle tension $f_i$ acting on the nonlinear elastic element and its extension $\Delta{n}_i$ can be expressed as an exponential function $f_i = e^{k_n \Delta{n}_i}$ (where $k_n$ is a constant) \cite{kawaharazuka2019musashi}.
  Therefore, it is possible to determine whether the muscle has been stretched as follows,
  \begin{align}
    \Delta{n}_{i, t+1}-\Delta{n}_{i, t} &> C_{src}\nonumber\\
    \frac{1}{k_n}\textrm{log}(f_{i, t+1})-\frac{1}{k_n}\textrm{log}(f_{i, t}) &> C_{src}\nonumber\\
    f_{i, t+1}-f_{i, t} &> C'_{src}f_{i, t}
  \end{align}
  where $C'_{src} = e^{k_n C_{src}}-1$.
  In other words, the stretch reflex should occur when the difference in muscle tension between the previous time step $\Delta{f}_i = f_{i, t+1} - f_{i, t}$ becomes larger than $C'_{src}f_{i, t}$.
  At this point, the target muscle length $l^{ref}_i$ is contracted by a fixed amount $\Delta{l}_{src, i}$.
  Afterward, over a time period $\Delta{t}_{src}$, the contracted length $\Delta{l}_{src, i}$ is gradually returned to its original state.
  This control not only offers the advantage of risk avoidance and posture stabilization to address Modeling Difficulty, but can also be actively applied to movements such as weightlifting.

  \subsubsection{Antagonist Inhibition Control (AIC)}
  Antagonist Inhibition Control (AIC) is a control method that mimics human reciprocal innervation, as mentioned earlier.
  It controls the suppression of the antagonist muscle relative to the agonist muscle.
  The muscle Jacobian $\bm{G}(\bm{\theta})$ is an $M \times N$ matrix that represents how much the muscles contract when the joint at angle $\bm{\theta}$ moves in a certain direction.
  In other words, $\bm{G}(\bm{\theta})(\bm{\theta}^{ref} - \bm{\theta})$ indicates whether each muscle acts as an agonist or an antagonist.
  In this method, the muscle stiffness $\bm{k}_{msc}$ of muscle stiffness control is modified depending on whether the muscle acts as an agonist or antagonist.
  For the $i$-th muscle, if $k_{msc, i}$ is 0, the muscle tension remains constant at $f^{bias}_{i}$, and if $k_{msc, i}$ is positive, it generates force in the direction that follows the target muscle length $l^{ref}_i$.
  The antagonist inhibition control is formulated as follows,
  \begin{align}
    \bm{s} &= \bm{G}(\bm{\theta})\frac{\bm{\theta}^{ref}-\bm{\theta}}{||\bm{\theta}^{ref}-\bm{\theta}||_{2}} \\
    k_{msc, i} &= k^{ref}_{msc, i} \;\;\;\; if \;\;\;\; s_i < C_{aic} \\
    k_{msc, i} &= 0 \;\;\;\;\;\;\;\;\;\; if \;\;\;\; s_i \geqq C_{aic}
  \end{align}
  where $\bm{s}$ represents the moment arm, $k^{ref}_{msc}$ is the constant applied to the stiffness of the agonist muscle, and $C_{aic}$ is the threshold that determines the antagonistic relationship.
  The default value of $C_{aic}$ is $0$, and muscle stiffness is adjusted based on the sign of the moment arm, indicating whether the muscle is an agonist or an antagonist.
  To ensure motion stability, it is also possible to set $C_{aic}$ to a positive value, which prevents the suppression of antagonist muscles with small moment arms.
  This control alleviates movement inhibition by Modeling Difficulty and High Internal Force of the musculoskeletal body, allowing for smooth movements across a wide range of motions.

  \subsubsection{Muscle Thermal Control (MTC)}
  Muscle Thermal Control (MTC) is a reflex mechanism where the activity of a muscle is suppressed as its temperature rises.
  Here, muscle tension is controlled according to the muscle temperature, which is represented here as the motor core temperature.
  This mechanism functions similarly to a tendon reflex.
  In this method, muscle length and muscle tension are not considered as vectors, and each muscle is treated individually.
  A two-resistor temperature model is adopted as the motor's temperature model, letting $C_{1}$ and $C_{2}$ be thermal capacities for the motor core and housing respectively, and $R_{1}$ and $R_{2}$ be thermal resistances between the motor core and motor housing, and between the motor housing and ambient air, respectively.
  Under this model, the motor core temperature $c_{1}$, motor housing temperature $c_{2}$, ambient temperature $c_{a}$, and muscle tension $f$ have the following relationship:
  \begin{align}
    \dot{c}_{1} &= \frac{K}{C_{1}}f^{2} - \frac{c_{1}-c_{2}}{R_{1}C_{1}} \label{eq:thermal-eq1}\\
    \dot{c}_{2} &= \frac{c_{1}-c_{2}}{R_{1}C_{2}} - \frac{c_{2}-c_{a}}{R_{2}C_{2}} \label{eq:thermal-eq2}
  \end{align}
  where $K$ is a constant calculated from the motor's torque constant, transmission efficiency, gear ratio, pulley radius, and winding resistance.
  By iterating the recurrence relations of \equref{eq:thermal-eq1} and \equref{eq:thermal-eq2} discretely, the motor core temperature $c_{1}$ can be estimated.
  Conversely, the time series of muscle tension $\bm{f}^{limit}_{seq}$ that can bring the motor core temperature $c_{1}$ to the set maximum value $c^{max}_{1}$ as quickly as possible can be calculated similarly to model predictive control.
  During this process, constraints on the minimum and maximum muscle tension and the smoothness of the time-series transition of muscle tension are added.
  To prevent the current muscle tension from exceeding the current value $f^{limit}_{t}$ of $\bm{f}^{limit}_{seq}$, the target muscle length $l^{ref}$ of muscle stiffness control is relaxed by $\Delta{l}_{mtc}$ as follows,
  \begin{align}
    &if\;\;f > f^{limit}_{t} \nonumber\\
    &\;\;\;\;\;\;\;\;\;\;\;\;\Delta{l}_{mtc, t} = \Delta{l}_{mtc, t-1} + \min(D_{gain}d-\Delta{l}_{mtc, t-1}, \Delta{l}_{plus}d)\nonumber\\
    &else \nonumber\\
    &\;\;\;\;\;\;\;\;\;\;\;\;\Delta{l}_{mtc, t} = \Delta{l}_{mtc, t-1} + \max(0-\Delta{l}_{mtc, t-1}, -\Delta{l}_{minus}d)\nonumber\\
    &d = |f-f^{limit}_{t}|&
  \end{align}
  where $|\cdot|$ denotes the absolute value, $\Delta{l}_{mtc, t}$ is the relaxation degree at time $t$, $\Delta{l}_{\{minus, plus\}}$ are coefficients that determine the muscle length change amount per step in the negative or positive direction, and $D_{gain}$ is the coefficient that determines the maximum relaxation amount.
  In other words, by imposing limits with $\Delta{l}_{minus}d$ and $\Delta{l}_{plus}d$, the muscle is relaxed or tightened to ensure the muscle tension does not exceed the maximum value.
  With this control, even if excessive muscle length is commanded or excessive force is applied, the muscles will automatically relax to ensure that the motor core temperature does not exceed $c^{max}_{1}$, thus addressing High Internal Force.
}%
{%
  \subsubsection{Stretch Reflex Control (SRC)}
  伸長反射制御\cite{kawaharazuka2020stretch}は先程述べたように人間の伸長反射を模倣した制御であり, 各筋$i$の筋長$l_i$が急に伸ばされたときに, その筋を収縮させる制御である.
  つまり, 筋長の変化$\Delta{l}_i=l_{i, t+1}-l_{i, t}$ ($l_{i, t}$は時刻$t$の筋長を表す)が, 閾値$C_{src}$ ($C_{src}$は定数)よりも大きくなったときに伸長反射が発生する.
  しかし, \secref{subsec:structure}で述べたように, 筋骨格ヒューマノイドには非線形弾性要素が含まれていることが多く, この場合衝撃で急に筋が伸ばされても, モータよりも柔軟で衝撃を受けることが可能な非線形弾性要素の変形が支配的となる.
  非線形弾性要素にかかる筋張力$f_i$とその伸び$\Delta{n}_i$の関係は指数関数$f_i = {e}^{k_n\Delta{n}_i}$ ($k_n$は定数とする)として表すことができるため\cite{kawaharazuka2019musashi}, 以下のように筋が伸ばされたかどうかを判定できる.
  \begin{align}
    \Delta{n}_{i, t+1}-\Delta{n}_{i, t} &> C_{src}\nonumber\\
    \frac{1}{k_n}\textrm{log}(f_{i, t+1})-\frac{1}{k_n}\textrm{log}(f_{i, t}) &> C_{src}\nonumber\\
    f_{i, t+1}-f_{i, t} &> C'_{src}f_{i, t}
  \end{align}
  ここで, $C'_{src}=e^{k_nC_{src}}-1$とする.
  つまり, 筋張力の前時刻との差分$\Delta{f}_i=f_{i, t+1}-f_{i, t}$が$C'_{src}$より大きくなったときに, 伸長反射を発生させれば良い.
  このとき, 指令筋長$l^{ref}_i$をある一定長さ$\Delta{l}_{src, i}$だけ収縮させる.
  その後, $\Delta{t}_{src}$時間かけて, 収縮された$\Delta{l}_{src, i}$分の筋を元に戻していく.
  この制御はModeling Difficultyに対処する危険回避や姿勢安定化の利点だけでなく, これを能動的に利用した重量挙げ等の動作に応用可能である.

  \subsubsection{Antagonist Inhibition Control (AIC)}
  拮抗筋抑制制御\cite{kawaharazuka2017antagonist}は先程述べたように人間の相反性神経支配を模倣した制御であり, 主動筋に対して拮抗筋を抑制するような制御である.
  筋長ヤコビアン$\bm{G}(\bm{\theta})$は, 関節角度が$\bm{\theta}$の際に, ある方向に関節を動かした際にどれだけ筋が縮むかを表す$M{\times}N$の行列である.
  つまり, $\bm{G}(\bm{\theta})(\bm{\theta}^{ref}-\bm{\theta})$はそれぞれの筋がある方向に動くときに主動筋となるか, 拮抗筋となるかを表現する.
  本手法では筋剛性制御の筋剛性$\bm{k}_{msc}$を主動筋か拮抗筋かによって変更する.
  $i$番目の筋について, $k_{msc, i}$は0であればその筋の張力は$f^{bias}_{i}$で一定となり, $k_{msc, i}$が正であれば指令筋長$l^{ref}_i$に追従する方向に力を発生させる.
  よってこれはまさに拮抗筋と主動筋であり, 拮抗筋抑制制御は以下のような決定をする.
  \begin{align}
    \bm{s} &= \bm{G}(\bm{\theta})\frac{\bm{\theta}^{ref}-\bm{\theta}}{||\bm{\theta}^{ref}-\bm{\theta}||_{2}} \\
    k_{msc, i} &= k^{ref}_{msc, i} \;\;\;\; if \;\;\;\; s_i < C_{aic} \\
    k_{msc, i} &= 0 \;\;\;\;\;\;\;\;\;\; if \;\;\;\; s_i \geqq C_{aic}
  \end{align}
  ここで, $\bm{s}$は$||\bm{\theta}^{ref}-\bm{\theta}||_{2}$で割ることでモーメントアームを表すようになり, $k^{ref}_{msc}$は主動筋の$k_{msc}$に与える定数, $C_{aic}$は拮抗関係を決める閾値である.
  $C_{aic}=0$が基本であり, モーメントアームの正負, つまり主動筋か拮抗筋かによって筋剛性を変更するが, 動作の安定性を確保するため$C_{aic}$を正に取ることで拮抗筋であるがモーメントアームの小さい筋を抑制しないようにすることも可能である.
  この制御はModeling Difficultyによる動きの阻害やHigh Internal Forceを緩和してスムーズに大きな可動域を動作することを可能にする.

  \subsubsection{Muscle Thermal Control (MTC)}
  筋温度制御\cite{kawaharazuka2020thermo}は, 筋肉の温度が上昇するとその筋肉の活動が抑制される反射機構であり, 筋肉の温度, ここではモータコアの温度に応じて筋張力を制御する.
  これはある種, 腱反射のようにも働いている.
  なお, ここでは筋長や筋張力をベクトルとして考えず, 各筋を個別に扱う.
  まず2抵抗温度モデルをモータの温度モデルとして採用し, モータコアとハウジングに熱容量$C_{1}, C_{2}$, モータコアとモータハウジング間, モータハウジングと外気温間に熱抵抗$R_{1}, R_{2}$を仮定する.
  このとき, モータコア温度$c_{1}$, モータハウジング温度$c_{2}$, 外気温$c_{a}$, 筋張力$f$の間には以下のような関係が存在する.
  \begin{align}
    \dot{c}_{1} &= \frac{K}{C_{1}}f^{2} - \frac{c_{1}-c_{2}}{R_{1}C_{1}} \label{eq:thermal-eq1}\\
    \dot{c}_{2} &= \frac{c_{1}-c_{2}}{R_{1}C_{2}} - \frac{c_{2}-c_{a}}{R_{2}C_{2}} \label{eq:thermal-eq2}
  \end{align}
  ここで, $K$はモータのトルク定数, 伝達効率, ギア比, プーリ半径, 巻線電気抵抗から計算される定数である.
  この\equref{eq:thermal-eq1}と\equref{eq:thermal-eq2}の漸化式を離散的に繰り返すことで, モータコア温度$c_{1}$を推定することができる.
  逆に, モータコア温度$c_{1}$を設定した最大値$c^{max}_{1}$に最速で達成させることが可能な筋張力の時系列$\bm{f}^{limit}_{seq}$は, モデル予測制御と同じ要領で計算することができる.
  なお, この際筋張力の最小値と最大値に関する制約, 筋張力の時系列遷移が滑らかであるという制約を加える.
  現在の筋張力が$\bm{f}^{limit}_{seq}$の現在時刻の値$f^{limit}_{t}$を越えないように, 筋剛性制御の指令筋長$l^{ref}$を$\Delta{l}_{mtc}$だけ弛緩させる.
  \begin{align}
    &if\;\;f > f^{limit}_{t} \nonumber\\
    &\;\;\;\;\;\;\;\;\;\;\;\;\Delta{l}_{mtc, t} = \Delta{l}_{mtc, t-1} + \min(D_{gain}d-\Delta{l}_{mtc, t-1}, \Delta{l}_{plus}d)\nonumber\\
    &else \nonumber\\
    &\;\;\;\;\;\;\;\;\;\;\;\;\Delta{l}_{mtc, t} = \Delta{l}_{mtc, t-1} + \max(0-\Delta{l}_{mtc, t-1}, -\Delta{l}_{minus}d)\nonumber\\
    &d = |f-f^{limit}_{t}|&
  \end{align}
  ここで, $|\cdot|$は絶対値, $\Delta{l}_{mtc, t}$は時刻$t$における弛緩度, $\Delta{l}_{\{minus, plus\}}$はマイナス方向またはプラス方向に対する一ステップの筋長変化量を決める係数, $D_{gain}$は最大弛緩量を決める係数である.
  つまり, $\Delta{l}_{minus}d$, $\Delta{l}_{plus}d$で制限をかけながら, 筋張力が最大値を越えないように筋を弛緩・緊張させている.
  この制御により, 無理な筋長が送られたり力がかかったりしても, モータコア温度が$c^{max}_{1}$を越えないように, 勝手に筋が弛緩するようになり, High Internal Forceに対処することができる.
}%

\begin{figure*}[t]
  \centering
  \includegraphics[width=1.9\columnwidth]{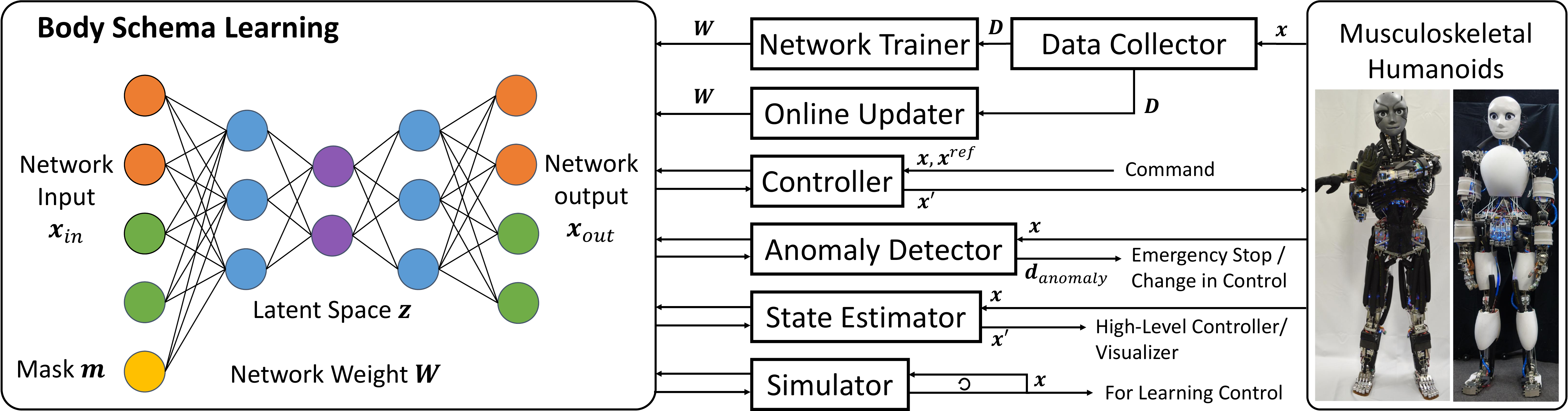}
  \caption{The overview of the body schema learning and control system. The system consists of the data collector, network trainer, online updater, controller, anomaly detector, state estimator, and simulator.}
  \label{figure:learning}
  \vspace{-1.0ex}
\end{figure*}

\switchlanguage%
{%
  \subsubsection{Maximum Speed Control}
  Maximum Speed Control is a method that adjusts the stiffness and length of muscles to enable rapid joint movements.
  Musculoskeletal humanoids have redundant muscles, each with different moment arms to the joints and varying maximum muscle length velocities.
  Muscles with larger moment arms can reach their maximum muscle length velocity more easily, even with the same joint angle velocity.
  This can impede movement and prevent the joint from achieving its intended angular velocity.
  Here, we describe two methods to solve this problem.

  The first method is Inhibition Control (IC), which suppresses antagonist muscles with large moment arms by setting the motor current to zero, thereby using backdrivability to rapidly release the muscles.
  This method is quite simple and sets the motor current of some muscles to zero based on the following equations,
  \begin{align}
    \bm{r} &= \bm{G}(\bm{\theta})\dot{\bm{\theta}}/||\dot{\bm{\theta}}||_{2} \\
    o_i &= 0\;\;\;\;if\;\;\;r_i/\dot{l}^{max} > C_{ic}
  \end{align}
  where $\bm{o}$ represents the motor current, and $C_{ic}$ is a constant.
  If $C_{ic} = 0$, the motor current of only the antagonist muscles becomes zero.
  If $C_{ic} > 0$, the motor current of antagonist muscles with large moment arms becomes zero.
  If the muscle actuators have backdrivability, the muscles will naturally extend when pulled, eliminating the maximum muscle length velocity for these muscles.

  The second method is Elongation Control (EC), which involves pre-elongating the antagonist muscles with large moment arms to enable rapid joint movements.
  This begins by simulating how pre-elongating antagonist muscles affects movement duration, or joint velocity.
  Let the current joint angle be $\bm{\theta}^{start}$ and the target joint angle be $\bm{\theta}^{end}$.
  A mask vector $\bm{m}$ is created, where muscles to be relaxed in advance are set to 0, and those not to be relaxed are set to 1.
  The following calculation is then performed,
  \begin{align}
    \underset{\Delta\bm{\theta}}{\textrm{minimize}}&\;\;\;(\bm{\theta}^{end}-\bm{\theta}-\Delta\bm{\theta})^{T}W_{3}(\bm{\theta}^{end}-\bm{\theta}-\Delta\bm{\theta})\label{eq:simulation}\\
    \textrm{subject to}&\;\;\; -\bm{m}\otimes\dot{\bm{l}}^{min}\Delta{t} \leq \bm{m}\otimes(G(\bm{\theta})\Delta\bm{\theta}) \leq \bm{m}\otimes\dot{\bm{l}}^{max}\Delta{t}\nonumber
  \end{align}
  where $\Delta\bm{\theta}$ represents the expected displacement from the current joint angle $\bm{\theta}$, $W_{3}$ is the weight matrix, and $\Delta{t}$ is the simulation interval.
  This allows the calculation of $\Delta\bm{\theta}$, which indicates how close the joint can get to $\bm{\theta}^{end}$ within $\Delta{t}$ seconds starting from $\bm{\theta}$.
  The simulation starts at $\bm{\theta} = \bm{\theta}^{start}$ and is updated iteratively as $\bm{\theta} \gets \bm{\theta} + \Delta\bm{\theta}$ and $t \gets t + \Delta{t}$ until $\bm{\theta}^{end}$ is reached.
  The total time taken is denoted as $\Delta{t}^{cost}$, and $\bm{m}$ is determined through exhaustive search to minimize $\Delta{t}^{cost}$.
  Finally, based on the muscle length transitions computed by \equref{eq:simulation}, the muscle length is pre-elongated to ensure that the muscles do not reach their maximum speed.
  This way, the antagonist muscles do not impede the agonist muscles, allowing for rapid movements through Free Joint Utilization.

  \subsubsection{Muscle Relaxation Control (MRC)}
  Muscle Relaxation Control is a simple control method that gradually relaxes muscles without affecting the posture.
  First, the necessary joint torque $\bm{\tau}^{nec}$ at the current joint angle $\bm{\theta}$ is calculated, and the muscle tension $\bm{f}^{nec}$ required to maintain this torque is computed using \equref{eq:tension-calc-2}.
  Next, the muscles are sorted in ascending order based on $\bm{f}^{nec}$.
  The muscles with the smallest $\bm{f}^{nec}$ values, i.e., those not essential for maintaining posture, are examined sequentially.
  If the current muscle tension sensor value $f_{i}$ for the muscle $i$ is less than $f^{min}$, the control moves on to the next muscle $i+1$.
  If $f_{i} > f^{min}$, the value of $\Delta{l}_{i}$ is checked.
  If $\Delta{l}_{i} + \Delta{l}_{+} > \Delta{l}^{max}$ ($\Delta{l}_{+}$ represents the muscle relaxation amount per step), the control proceeds to the next muscle $i+1$.
  If $\Delta{l}_{i} + \Delta{l}_{+} < \Delta{l}^{max}$, then $\Delta{l}_{i}$ is updated as $\Delta{l}_{i} \gets \textrm{min}(\Delta{l}_{i} + \Delta{l}_{+}, \Delta{l}^{max})$.
  If the update of $\Delta{l}_{i}$ succeeds even once, the loop exits.
  If not, the process is repeated for all muscles.

  This describes the basic operation of muscle relaxation control, but several conditions apply to its activation.
  Firstly, muscle relaxation control operates only in a static state where the muscle length command $\bm{l}^{ref}$ is not fluctuating.
  When $\bm{l}^{ref}$ is fluctuating, i.e., during movement, the muscles with the highest $\bm{f}^{nec}$ are targeted in descending order, and $\Delta{l}_{i}$ is reverted to $0$ as $\Delta{l}_{i} \gets \textrm{max}(\Delta{l}_{i} - \Delta{l}_{-}, 0)$ ($\Delta{l}_{-}$ represents the muscle contraction amount per step).
  Additionally, muscle relaxation control stops if the norm of the difference between the current joint angle $\bm{\theta}$ and the joint angle $\bm{\theta}^{init}$ at the moment when $\bm{l}^{ref}$ stopped is greater than a maximum allowable change $\Delta{\theta}^{max}$.
  This ensures that muscle relaxation does not significantly affect the posture.
  This control allows for the suppression of High Internal Force in antagonistic relationships, which enables continuous movement, internal force suppression due to contact with the environment, and actions such as using the environment to rest the body.
}%
{%
  \subsubsection{Maximum Speed Control}
  最大速度制御\cite{kawaharazuka2020speed}は, 関節の素早い動作を可能にするために筋肉の硬さや長さを変化させる制御である.
  筋骨格ヒューマノイドには冗長な筋肉が存在し, それらの筋肉の関節に対するモーメントアームは様々で, かつ最大筋長速度も異なる.
  モーメントアームが大きな筋肉は, 同じ関節角速度でも筋長速度が大きくなり, 最大筋長速度に達しやすく, それらが動きの阻害となって本来出せるはずの関節角度がでなくなってしまう場合が多々ある.
  ここでは, この問題を解決する2つの方法について述べる.

  一つ目はInhibition Controlであり, モーメントアームの大きな拮抗筋を抑制, つまり流れる電流を0とすることで, バックドライバビリティを使って筋を素早く繰り出すという手法である.
  これは非常に単純で, 以下の式に基づき一部の筋の電流値を0にする.
  \begin{align}
    \bm{r} &= \bm{G}(\bm{\theta})\dot{\bm{\theta}}/||\dot{\bm{\theta}}||_{2} \\
    o_i &= 0\;\;\;\;if\;\;\;r_i/\dot{l}^{max} > C_{ic}
  \end{align}
  ここで, $\bm{o}$は電流値, $C_{ic}$は定数である.
  $C_{ic}=0$とすると, 拮抗筋のみの電流値が0となり, $C_{ic}>0$とすると, 拮抗筋の中でもモーメントアームが大きな筋のみの電流値が0となる.
  もし筋アクチュエータにバックドライバビリティがあれば, 筋は引っ張られることで自然と筋が伸びていき, それらの筋の最大筋長速度が存在しなくなる.

  二つ目はElongation Controlであり, モーメントアームの大きな拮抗筋を予め緩めておくという方法である.
  まず, 拮抗筋を緩めることで動作時間, つまり関節速度がどう変化するかをシミュレートする.
  動作初期の関節角度を$\bm{\theta}^{start}$, 指令関節角度を$\bm{\theta}^{end}$とする.
  予め緩める筋を0, 緩めない筋を1としたマスクベクトル$\bm{m}$を決め, 以下の計算を行う.
  \begin{align}
    \underset{\Delta\bm{\theta}}{\textrm{minimize}}&\;\;\;(\bm{\theta}^{end}-\bm{\theta}-\Delta\bm{\theta})^{T}W_{3}(\bm{\theta}^{end}-\bm{\theta}-\Delta\bm{\theta})\label{eq:simulation}\\
    \textrm{subject to}&\;\;\; -\bm{m}\otimes\dot{\bm{l}}^{limit}\Delta{t} \leq \bm{m}\otimes(G(\bm{\theta})\Delta\bm{\theta}) \leq \bm{m}\otimes\dot{\bm{l}}^{limit}\Delta{t}\nonumber
  \end{align}
  ここで, $\Delta\bm{\theta}$は現在関節角度$\bm{\theta}$からの予想される変位, $W_{3}$は重み行列, $\Delta{t}$はシミュレーションの間隔を表す.
  これにより, $\bm{\theta}$から$\Delta{t}$秒間で最大どの程度$\bm{\theta}^{end}$に近づけるかを表す$\Delta\bm{\theta}$を計算することができる.
  このシミュレーションを$\bm{\theta}=\bm{\theta}^{start}$からはじめ, $\bm{\theta}\gets\bm{\theta}+\Delta\bm{\theta}$, $t{\gets}t+\Delta{t}$というように$\bm{\theta}^{end}$になるまで更新していく.
  このときにかかった時間を$\Delta{t}^{cost}$とし, $\Delta{t}^{cost}$が小さくなるような$\bm{m}$を全探索によって求める.
  最後に\equref{eq:simulation}により計算可能な筋長の遷移から, 最大速度に達しないよう予め筋長を伸ばしておく.
  これにより, 拮抗筋が主動筋を阻害せず, Free Joint Utilizationにより, 素早い動作が可能となる.

  \subsubsection{Muscle Relaxation Control (MRC)}
  筋弛緩制御\cite{kawaharazuka2019relax}は, 姿勢に影響を及ぼさない範囲で徐々に筋を弛緩させていくシンプルな制御である.
  まず, 現在の関節角度$\bm{\theta}$において必要な関節トルクを$\bm{\tau}^{nec}$として, \equref{eq:tension-calc-2}から現在必要と考えられる筋張力$\bm{f}^{nec}$を計算する.
  次に, $\bm{f}^{nec}$の値から筋を昇順にソートし, $\bm{f}^{nec}$の小さい, つまり姿勢の維持に必要のない筋$i$から順に見ていく.
  現在の実機の筋張力センサ値$f_{i}$が$f_{i}<f^{min}$の場合は次の筋$i+1$へ移動する.
  もし$f_{i}>f^{min}$の場合は, $\Delta{l}_{i}$の値を確認する.
  もし$\Delta{l}_{i}+\Delta{l}_{+}>\Delta{l}^{max}$ならば, 次の筋$i+1$へ移動する($\Delta{l}_{+}$は1ステップにおける筋の弛緩量を表す).
  $\Delta{l}_{i}+\Delta{l}_{+}<\Delta{l}^{max}$ならば, $\Delta{l}_{i} \gets \textrm{min}(\Delta{l}_{i} + \Delta{l}_{+}, \Delta{l}^{max})$とし, $\Delta{l}_{i}$を更新する.
  一度でも$\Delta{l}_{i}$の更新に成功すればループを抜け, 成功しない場合は最後の筋まで上記の工程を繰り返す.

  これが基本的な筋弛緩制御の動きであるが, その作動に関してはいくつかの条件が入る.
  まず, 筋弛緩制御は筋長指令$\bm{l}^{ref}$が変動しない, つまり静止状態においてのみ動作する.
  $\bm{l}^{ref}$が変動している, つまり動作中に関しては, 逆に$\bm{f}^{nec}$が高い筋から順に, $\Delta{l}_{i} \gets \textrm{max}(\Delta{l}_{i}-\Delta{l}_{-}, 0)$のように, $\Delta{l}_{i}$を$0$に戻していく($\Delta{l}_{-}$は1ステップにおける筋の収縮量を表す).
  また, $\bm{l}^{ref}$が静止した瞬間の関節角度を$\bm{\theta}^{init}$, 現在の関節角度を$\bm{\theta}$としたときに, $||\bm{\theta}-\bm{\theta}^{init}||_{2}>\Delta{\theta}^{max}$となった場合には筋弛緩制御を停止する($\Delta{\theta}^{max}$は許容可能な関節角度の最大変化量を表す).
  これは, 筋の弛緩によって姿勢に大きな影響を及ぼさないようにするためである.
  これにより, 拮抗関係におけるHigh Internal Forceを抑え継続的な動作が可能となると同時に, 環境との接触による内力の抑制, そして環境を使って身体を休める動作等が可能となる.
}%

\subsection{Body Schema Learning and Control for Musculoskeletal Humanoids} \label{subsec:learning}
\switchlanguage%
{%
  We describe the body schema learning, the layer above the reflex control, as well as control and state estimation using this schema.
  In this study, we refer to a model that represents the relationships between sensory input and motor output, describing correlations such as how movements of the body affect vision, touch, sound, and temperature, as a body schema \cite{head1911bodyschema, hoffmann2010bodychema}.
  The overall system including the body schema is shown in \figref{figure:learning}.
  The body schema in this study is represented as follows,
  \begin{align}
    \bm{z} &= \bm{h}_{enc}(\bm{x}_{in}, \bm{m})\\
    \bm{x}_{out} &= \bm{h}_{dec}(\bm{z})\\
    \bm{x}_{out} &= \bm{h}(\bm{x}_{in}, \bm{m})
  \end{align}
  where $\bm{x}_{in}$ is the network input, $\bm{x}_{out}$ is the network output, $\bm{m}$ is the mask variable, $\bm{z}$ is the latent variable, $\bm{h}_{enc}$ represents the encoder part of the neural network, $\bm{h}_{dec}$ represents the decoder part, and $\bm{h}$ represents the entire network.
  The mask variable is used to represent the correlation between the sensors included in $\bm{x}_{\{in, out\}}$ by masking some of the inputs to predict the outputs.
  The variables $\bm{x}_{\{in, out\}}$ are consolidated and generalized as $\bm{x}$, which includes various sensor values and control inputs such as joint angle, muscle tension, muscle length, and contact force.
  Using this body schema as the core, the system can perform data collection, network training, and online update, as well as control, state estimation, anomaly detection, and simulation for the musculoskeletal humanoid.
  When the time at which $\bm{x}_{out}$ is obtained is the same as the time at which $\bm{x}_{in}$ is obtained, it is referred to as a static body schema.
  When they differ, i.e., $\bm{x}_{out}$ is at time $t+1$ and $\bm{x}_{in}$ is at time $t$, it is called a dynamic body schema.
  The static body schema \cite{kawaharazuka2024bodyschema} represents the correlation between sensor values at the same time, while the dynamic body schema \cite{kawaharazuka2023dpmpb} represents the temporal state transitions of sensor values.
  We will summarize the system for musculoskeletal humanoids for each of these schemas.

  \subsubsection{Static Body Schema}
  The static body schema in the musculoskeletal structure represents the correlations among various sensors, particularly the joint angle $\bm{\theta}$, muscle tension $\bm{f}$, and muscle length $\bm{l}$, which are indispensable in musculoskeletal systems.
  This is also referred to as joint-muscle mapping.
  There are three relationships among these three sensors: it is possible to estimate $\bm{l}$ from $\bm{\theta}$ and $\bm{f}$, estimate $\bm{\theta}$ from $\bm{f}$ and $\bm{l}$, and estimate $\bm{f}$ from $\bm{l}$ and $\bm{\theta}$.
  The variables are set as follows.
  \begin{align}
    \bm{x}^T_{out} &= \bm{x}^T_{in} = \begin{pmatrix}\bm{\theta}^T & \bm{f}^T & \bm{l}^T\end{pmatrix}\\
    \bm{m} &= \{\begin{pmatrix}1 & 1 & 0\end{pmatrix}, \begin{pmatrix}1 & 0 & 1\end{pmatrix}, \begin{pmatrix}0 & 1 & 1\end{pmatrix}\}
  \end{align}
  The input and output of the network are $\bm{\theta}$, $\bm{f}$, and $\bm{l}$, and the mask variable $\bm{m}$ represents their respective relationships.
  For example, when $\bm{m}$ is $\begin{pmatrix}1 & 1 & 0\end{pmatrix}$, $\bm{l}$ is treated as $\bm{0}$, and $\bm{l}$ is estimated from $\bm{\theta}$ and $\bm{f}$.
  In other words, all of $\bm{\theta}$, $\bm{f}$, and $\bm{l}$ are estimated from $\bm{\theta}$ and $\bm{f}$.
  After collecting data $\bm{x}$ from the actual robot, a static body schema of the musculoskeletal structure can be learned by randomly varying $\bm{m}$, masking $\bm{x}$, and training the network accordingly.
  To cope with temporal changes in the body schema, online learning is also possible, in which the network weights $\bm{W}$ are updated in real time based on the data $\bm{x}$.
  If we want to find the muscle length that minimizes muscle tension while bringing the joint angle closer to the target value $\bm{\theta}^{ref}$, the network can be used to optimize $\bm{z}$ through forward and backward propagation to minimize $||\bm{\theta}^{ref} - \bm{\theta}||_{2} + ||\bm{f}||_{2}$, ultimately obtaining $\bm{l}$.
  In a similar manner, it is also possible to estimate joint angles from muscle length and muscle tension, simulate joint angle and muscle tension changes based on muscle length changes, and detect anomalies using the reconstruction error of network outputs for muscle length and muscle tension.
  The static body schema training can overcome Modeling Difficulty.
  Additionally, by using this static body schema to realize the desired muscle tensions, it becomes possible to achieve Variable Stiffness Control, allowing impacts to be either softly absorbed or rigidly resisted as needed.

  Note that the simpler the network structure, the easier it is to train the body schema.
  If the goal is simply to obtain muscle length from the target joint angle, $\bm{x}_{in} = \bm{\theta}$ and $\bm{x}_{out} = \bm{l}$ can be set accordingly \cite{kawaharazuka2018online}.
  Furthermore, by setting $\bm{x}^T_{in} = \begin{pmatrix}\bm{\theta}^T & \bm{f}^T\end{pmatrix}$ and $\bm{x}^T_{out} = \bm{l}^T$, control that takes muscle tension into account becomes possible \cite{kawaharazuka2018bodyimage, kawaharazuka2019longtime}.

  \subsubsection{Dynamic Body Schema}
  The dynamic body schema in the musculoskeletal structure is a state equation that represents the time-series changes of various sensor values, including joint angle, muscle tension, muscle length, muscle temperature, and contact force.
  Here, we will explain object grasping with a musculoskeletal hand as an example of the dynamic body schema.

  The musculoskeletal hand of Musashi \cite{makino2018hand} is equipped with sensors that measure the wrist joint angle $\bm{\theta}$, muscle tension $\bm{f}$, muscle length $\bm{l}$, and contact force $\bm{f}_{contact}$ at the fingertips and palm.
  The dynamic body schema that expresses the time-series changes in each sensor value when an object is grasped and the fingers are moved is represented as follows,
  \begin{align}
    \bm{x}^T_{in} &= \begin{pmatrix}\bm{\theta}^T_t & \bm{f}^T_t & \bm{l}^T_t & \bm{f}^T_{contact, t} & \Delta\bm{l}^{ref}\end{pmatrix}\\
    \bm{x}^T_{out} &= \begin{pmatrix}\bm{\theta}^T_{t+1} & \bm{f}^T_{t+1} & \bm{l}^T_{t+1} & \bm{f}^T_{contact, t+1}\end{pmatrix}\\
    \bm{m} &= \{\begin{pmatrix}1 & 1 & 1 & 1 & 1\end{pmatrix}, \begin{pmatrix}0 & 0 & 0 & 0 & 1\end{pmatrix}\}
  \end{align}
  where $\Delta\bm{l}^{ref}$ is the control input, which is the difference in the muscle length command.
  The mask variable $\bm{m}$ indicates that the changes in sensor values at time $t+1$ can be inferred from the sensor values and control inputs at time $t$.
  Moreover, if this network $\bm{h}$ is a recurrent neural network that can take time series information into account, it can predict the sensor values at the next time step using only the control inputs.
  By collecting time-series data $\bm{x}$ from the actual robot while randomly commanding $\Delta\bm{l}^{ref}$, and training the network while randomly varying $\bm{m}$, it is possible to learn a dynamic body schema of the musculoskeletal hand.
  To cope with temporal changes in the body schema, online learning is also possible, in which the network weights $\bm{W}$ or learnable latent variables \cite{tani2002parametric} are updated in real time based on the data $\bm{x}$.
  If it is necessary to adjust the control inputs to maintain the initial contact force $\bm{f}^{init}_{contact}$ consistently throughout some movement, $\Delta\bm{l}^{ref}$ can be optimized by iteratively performing forward and backward propagation to minimize $||\bm{f}^{init}_{contact} - \bm{f}_{contact}||_{2}$ in the same manner as model predictive control.
  As this is a state equation, not only contact simulation and estimation but also anomaly detection using reconstruction errors $d_{anomaly}$ are possible.
  Additionally, although not detailed here, by using Parametric Bias \cite{tani2002parametric}, which can represent the time-series changes of various sensor values when grasping different objects within a single network, it is even possible to recognize what object is currently being grasped.
  These methods are applicable not only to the musculoskeletal hand but also to various body parts and object manipulation scenarios \cite{kawaharazuka2020dynamics, kawaharazuka2022cloth, kawaharazuka2021imitation}.
}%
{%
  反射制御よりも上位の身体図式学習とそれを用いた制御や状態推定について述べる.
  本研究では, 感覚入力や制御入力の間の関係性を表現し, 体をどう動かすと視界や接触, 音や温度がどう変化するといった相関を記述するモデルを身体図式\cite{head1911bodyschema, hoffmann2010bodychema}と呼んでいる.
  この身体図式を含む全体システムを\figref{figure:learning}に示す.
  本研究の身体図式は以下のように表される.
  \begin{align}
    \bm{z} &= \bm{h}_{enc}(\bm{x}_{in}, \bm{m})\\
    \bm{x}_{out} &= \bm{h}_{dec}(\bm{z})\\
    \bm{x}_{out} &= \bm{h}(\bm{x}_{in}, \bm{m})
  \end{align}
  ここで, $\bm{x}_{in}$はネットワーク入力, $\bm{x}_{out}$はネットワーク出力, $\bm{m}$はマスク変数, $\bm{z}$は潜在変数, $\bm{h}_{enc}$はニューラルネットワークで表現されたエンコーダ部, $\bm{h}_{dec}$はデコーダ部, $\bm{h}$はそれらを合わせた全体ネットワークを表す.
  マスク変数とは, $\bm{x}_{\{in, out\}}$に含まれるセンサ同士の相関関係を表現するために, 一部の入力をマスクして出力を予測するための変数である.
  $\bm{x}_{\{in, out\}}$をまとめ一般化した変数を$\bm{x}$とし, これは関節角度や筋張力, 筋長, 接触力などの多様なセンサ値・制御入力を含んでいる.
  この身体図式を核として, データ収集, ネットワーク学習, ネットワークのオンライン更新に加え, 筋骨格ヒューマノイドの制御, 状態推定, 異常検出, シミュレーションが可能である.
  このとき, $\bm{x}_{out}$を得た時間と$\bm{x}_{in}$を得た時間が同一の場合を静的身体図式, 異なる, つまり$\bm{x}_{out}$が$t+1$, $\bm{x}_{in}$が$t$の場合を動的身体図式と呼ぶ.
  静的身体図式\cite{kawaharazuka2024bodyschema}は同時刻におけるセンサ値間の相関関係を表し, 動的身体図式\cite{kawaharazuka2023dpmpb}は時間的なセンサ値の状態遷移を表す.
  それぞれについて筋骨格ヒューマノイドにおけるシステムをまとめる.

  \subsubsection{Static Body Schema}
  筋骨格構造における静的身体図式は, 各センサ間, 特に筋骨格系には欠かせない関節角度$\bm{\theta}$, 筋張力$\bm{f}$, 筋長$\bm{l}$の間の相関関係を表す\cite{kawaharazuka2020autoencoder}.
  これは関節-筋空間マッピングとも呼ばれる.
  この3つのセンサには3つの関係があり, $\bm{\theta}$と$\bm{f}$から$\bm{l}$を, $\bm{f}$と$\bm{l}$から$\bm{\theta}$を, $\bm{l}$と$\bm{\theta}$から$\bm{f}$を推定することができる.
  つまり, 以下のように変数を設定する.
  \begin{align}
    \bm{x}^T_{out} &= \bm{x}^T_{in} = \begin{pmatrix}\bm{\theta}^T & \bm{f}^T & \bm{l}^T\end{pmatrix}\\
    \bm{m} &= \{\begin{pmatrix}1 & 1 & 0\end{pmatrix}, \begin{pmatrix}1 & 0 & 1\end{pmatrix}, \begin{pmatrix}0 & 1 & 1\end{pmatrix}\}
  \end{align}
  ネットワークの入出力はともに$\bm{\theta}$, $\bm{f}$, $\bm{l}$であり, マスク変数$\bm{m}$によってそれぞれの関係を表現する.
  例えば$\bm{m}$が$\begin{pmatrix}1 & 1 & 0\end{pmatrix}$の場合, $\bm{l}=\bm{0}$として, $\bm{\theta}$と$\bm{f}$から$\bm{l}$を推定, つまり$\bm{\theta}$と$\bm{f}$から$\bm{\theta}$, $\bm{f}$, $\bm{l}$の全てを推定する.
  実機に置いて$\bm{x}$のデータを収集した後, $\bm{m}$をランダムに変化させ$\bm{x}$をマスクしつつ, ネットワークを学習させることで, 筋骨格構造における静的身体図式を学習することが可能である.
  逐次的な身体図式の変化に対応するために, $\bm{x}$のデータからリアルタイムにネットワークの重み$\bm{W}$を更新するオンライン学習も可能である.
  例えば関節角度を指令値$\bm{\theta}^{ref}$に近づけつつ筋張力を最小化する筋長を求めたい場合, $||\bm{\theta}^{ref}-\bm{\theta}||_{2}+||\bm{f}||_{2}$を最小化するように順伝播と誤差逆伝播を繰り返し$\bm{z}$を最適化, 最終的に$\bm{l}$を得ることができる.
  同様の形で筋長と筋張力からの関節角度推定や, 筋長変化からの関節角度と筋張力制御変化のシミュレーション, また筋長と筋張力のネットワーク出力の復元誤差を用いた異常検知も可能である.
  これらのように, ニューラルネットワークの逐次的な学習によってModeling Difficultyを克服することができる.
  また, この静的身体図式を用いて所望の筋張力を実現することで, ハードウェアによるVariable Stiffnessを実現することが可能であり, 衝撃を柔らかくいなしたり, 固く受け止めたりすることができる.
  なお, 単に指令関節角度から筋長を求めたいだけであれば, $\bm{x}_{in}=\bm{\theta}$, $\bm{x}_{out}=\bm{l}$のように設定しても良い\cite{kawaharazuka2018online}.
  他にも, $\bm{x}^T_{in}=\begin{pmatrix}\bm{\theta}^T & \bm{f}^T\end{pmatrix}$, $\bm{x}^T_{out}=\bm{l}^T$のように設定することで, 筋張力を考慮した制御が可能である\cite{kawaharazuka2018bodyimage, kawaharazuka2019longtime}.

  \subsubsection{Dynamic Body Schema}
  筋骨格構造における動的身体図式は, 関節角度や筋張力, 筋長, 筋温度や接触センサなど, 多様なセンサ値の時系列変化を表現する状態方程式である.
  これには様々なバリエーションが考えられ, 筋骨格ハンドの物体把持\cite{kawaharazuka2020dynamics}, 筋骨格身体による柔軟物体操作\cite{kawaharazuka2022cloth}, 筋骨格身体の摸倣学習\cite{kawaharazuka2021imitation}などが挙げられる.
  ここでは筋骨格ハンドの物体把持を例に説明する\cite{kawaharazuka2020dynamics}.
  Musashiに備わる筋骨格ハンド\cite{makino2018hand}には, 手首の関節角度$\bm{\theta}$, 筋張力$\bm{f}$, 筋長$\bm{l}$, 指先と手のひらの接触力$\bm{f}_{contact}$を測るセンサが搭載されている.
  何らかの物体を把持し指を動かしたときの各センサ値の時系列変化を表現する動的身体図式は, 以下のように表される.
  \begin{align}
    \bm{x}^T_{in} &= \begin{pmatrix}\bm{\theta}^T_t & \bm{f}^T_t & \bm{l}^T_t & \bm{f}^T_{contact, t} & \Delta\bm{l}^{ref}\end{pmatrix}\\
    \bm{x}^T_{out} &= \begin{pmatrix}\bm{\theta}^T_{t+1} & \bm{f}^T_{t+1} & \bm{l}^T_{t+1} & \bm{f}^T_{contact, t+1}\end{pmatrix}\\
    \bm{m} &= \{\begin{pmatrix}1 & 1 & 1 & 1 & 1\end{pmatrix}, \begin{pmatrix}0 & 0 & 0 & 0 & 1\end{pmatrix}\}
  \end{align}
  ここで, $\Delta\bm{l}^{ref}$は制御入力である筋長指令の差分である.
  この$\bm{m}$は, 時刻$t$のセンサ値と制御入力から次時刻$t+1$のセンサ値の変化がわかること, また, もしこのネットワーク$\bm{h}$が時系列を考慮可能なリカレントニューラルネットワークであれば, 制御入力のみから次時刻のセンサ値を予測することが可能であることを表している.
  $\Delta\bm{l}^{ref}$をランダムに指令した際の実機のデータ$\bm{x}$を時系列に収集し, $\bm{m}$をランダムに変化させつつネットワークを学習させることで, 筋骨格ハンドにおける動的身体図式を学習することが可能である.
  逐次的な身体図式の変化に対応するために, $\bm{x}$のデータからリアルタイムにネットワークの重み$\bm{W}$または学習可能な潜在変数\cite{tani2002parametric}を更新するオンライン学習も可能である.
  例えば何らかの動作中に, 常に最初に物体を把持したときの接触力$\bm{f}^{init}_{contact}$を保持し続けるように制御入力を調整したい場合, モデル予測制御と同じ要領で$||\bm{f}^{init}_{contact}-\bm{f}_{contact}||_{2}$を最小化するように順伝播と誤差逆伝播を繰り返し$\Delta\bm{l}^{ref}$を最適化すれば良い.
  また, これは状態方程式のため, 接触のシミュレーションや推定はもちろんのこと, 復元誤差から異常検知も可能である.
  加えてここでは詳しくは述べないが, 様々な物体を把持した際の多様なセンサ値の時系列変化を一つのネットワークで表現可能なParametric Bias \cite{tani2002parametric}を用いることで, 現在何を把持しているかという把持物体認識まで可能である.
  これらは筋骨格ハンドのみならず, 様々な身体部位や物体操作にも応用可能である.
}%

\begin{figure}[t]
  \centering
  \includegraphics[width=0.95\columnwidth]{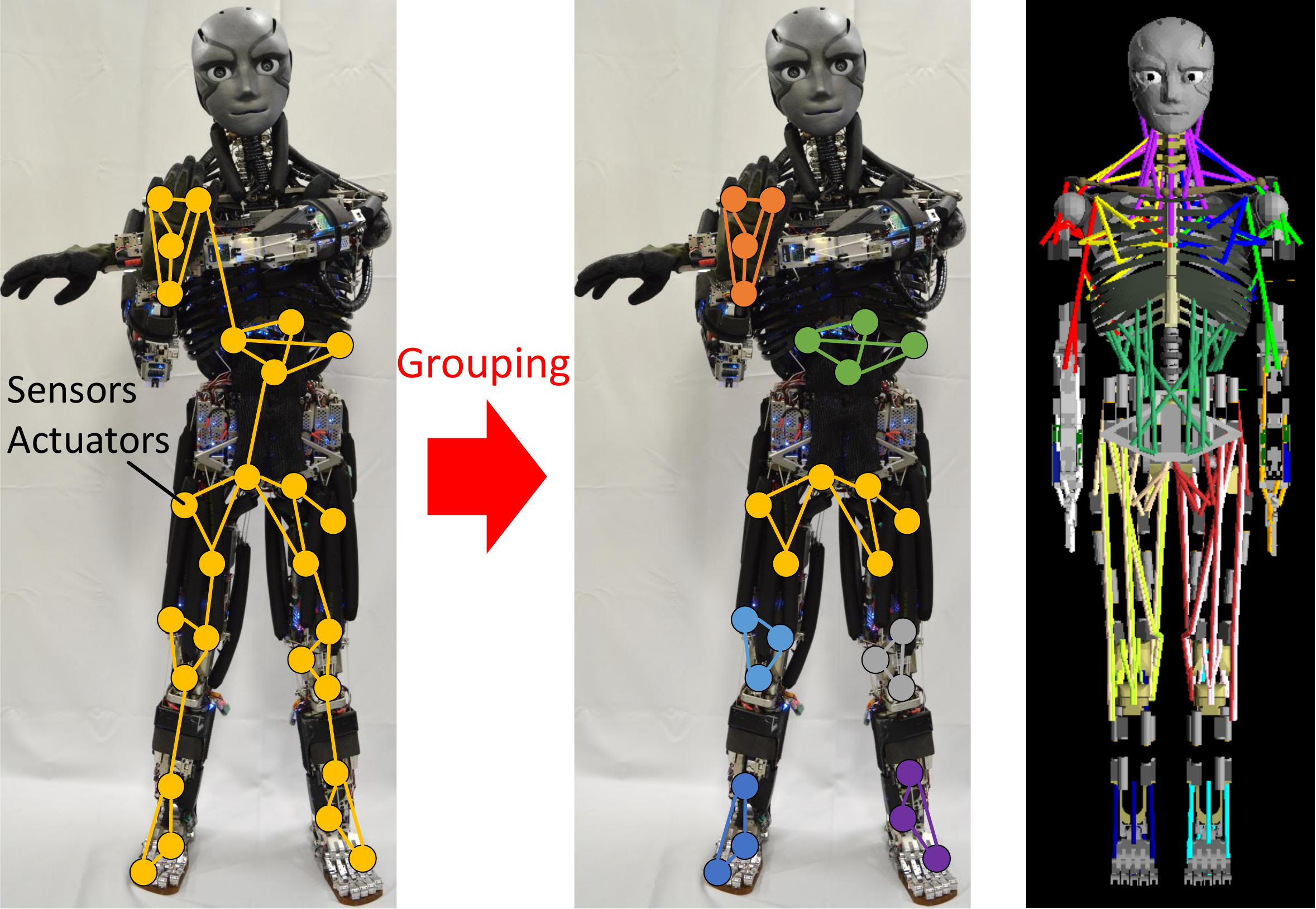}
  \caption{The grouping of muscles in body schema of musculoskeletal humanoids for reduction of high computational cost.}
  \label{figure:grouping}
  \vspace{-1.0ex}
\end{figure}

\subsection{Muscle Grouping for Musculoskeletal Humanoids} \label{subsec:grouping}
\switchlanguage%
{%
  Controlling all muscles simultaneously in a musculoskeletal humanoid is computationally intensive and impractical.
  To reduce this high computational cost, it is necessary to perform muscle grouping for the body schema.
  As shown in \figref{figure:grouping}, multiple muscles are arranged around each joint in the musculoskeletal humanoid.
  By appropriately grouping these muscles, the static body schema can be divided, thereby reducing the computational cost.
  There are two methods for this: manual grouping \cite{kawaharazuka2018estimator} and automatic grouping \cite{kawaharazuka2021grouping}, which are described below.

  Manual Grouping reduces computational cost by manually partitioning the static body schema into groups of joints and muscles.
  This method is straightforward: a small number of target joints for which the body schema is to be constructed are selected, and the muscle groups capable of moving those joints are identified.
  The joints that these muscles can move are then added to the target joint group, forming a single group.
  For example, to estimate joint angle of the 3 degrees of freedom of the shoulder joint, the muscle group capable of moving these 3 degrees of freedom is selected.
  This muscle group includes several multi-articular muscles such as the biceps brachii and pectoralis major, which also influence other joints.
  Consequently, joints like the elbow and scapula are included in the group.
  In the case of the shoulder of Kengoro, for instance, the static body schema corresponds to 10 muscles and 10 degrees of freedom.
  Similarly, constructing a static body schema for the neck requires a correspondence of 10 muscles and 16 degrees of freedom.
  Joint angle estimation is performed only on the initially targeted joints, while values estimated from other static body schemas are used for the remaining joints.
  Although various granularities can be considered for partition sizes, it is generally preferable, from a maintenance perspective, to have a small number of large static body schemas rather than many small ones.

  Automatic Grouping automates the manual grouping process to achieve more appropriate groupings.
  Two types of information aid in grouping the numerous redundant muscles spread throughout the body: functional connections and spatial connections.
  Functional connections indicate that muscles, due to their redundancy, are functionally related to each other, reflecting the strength of their correlations (e.g., the relationship between agonist and antagonist muscles is strong).
  Spatial connections represent the spatial proximity of muscles based on their neural connections (e.g., the spatial connection between leg muscles and arm muscles is weak).
  The former is derived from the network weights of the static body schema, while the latter is calculated from the geometric model and embedded into a graph structure. 
  By performing graph partitioning, appropriate muscle grouping is achieved automatically.
  Through this method, the robot can identify functional connections from its random movements and combine these with spatial connections to achieve appropriate muscle grouping.
  This allows for the creation of interpretable and manageable controllers for each group, maintaining accuracy while reducing computational cost and complexity.
}%
{%
  全ての筋肉を同時に制御することは計算量が非常に高く, 実用的ではない.
  この高い計算コストを削減するために, 身体図式における筋のグルーピングを行う必要がある.
  \figref{figure:grouping}に示すように, 筋骨格ヒューマノイドの各関節には複数の筋が配置されており, これらの筋を適切にグルーピングすることで静的身体図式を分割し, 計算量を削減することができる.
  この方法には手動グルーピングと自動グルーピングがあり, それぞれについて述べる.

  手動グルーピング\cite{kawaharazuka2018estimator}では, 静的身体図式をいくつかの関節と筋のグループに小分けることで計算量を削減する.
  方法は非常に単純で, 身体図式を構築したい, ターゲットとなる関節群を少数取り出し, その関節群を動作させることのできる筋群を得る.
  そして, その筋群が動かすことのできる関節群を取り出し, これをターゲットとなる関節群に加え一つのグループとする.
  例を挙げると, まず、肩の3自由度の関節角度推定を行いたい場合には, その3自由度を動かすことのできる筋群を取り出す.
  その筋群には上腕二頭筋や大胸筋等の多関節が多数含まれており, その筋群が動かすことのできる関節群をさらに取り出す.
  これには, 肘関節や肩甲骨が含まれており, 例えばKengoroの場合には, 最終的に10本の筋群と10自由度の関節群の対応である静的身体図式が構成される.
  同様の理論で首の静的身体図式作成には16自由度と10本の筋の対応が必要である.
  関節角度推定は最初にターゲットとした関節のみに対して行い, その他の関節については, 別の静的身体図式から推定された値を併用して用いる.
  分割サイズは様々な粒度が考えられるが, 小さな静的身体図式が多いよりは、保守管理の観点から、計算できるギリギリの大きさの静的身体図式が少数あったほうが望ましい.

  自動グルーピング\cite{kawaharazuka2021grouping}では, 手動で行ったグルーピングを自動化し, より適切なグルーピングを行う.
  全身に配置された多数の冗長な筋群をグルーピングするときに助けとなる情報には, 機能的接続と空間的接続がある.
  機能的接続とは, 筋は冗長であるがゆえにそれぞれが機能的に関係しあっており, その相関の強さを表す(e.g. 主動筋と拮抗筋の関係性は強い).
  空間的接続とは, 神経的な接続から来る筋同士の空間的な近さを表す(e.g. 足の筋と腕の筋の空間的な接続は弱い).
  前者の情報を静的身体図式のネットワーク重みから, 後者の情報を幾何モデルから計算してグラフ構造に埋め込み, グラフ分割を行うことで, 適切な筋グルーピングを自動的に行う.
  これにより, ロボットは自身のランダムな動きから機能的接続を見出し, これと空間的接続を合わせることで適切な筋グルーピングを行うことができる.
  また, 計算量や複雑性を下げつつも正確性を保った解釈性が高く扱いやすい制御器をグルーピングごとに自動的に作り上げていくことが可能となる.
}%

\begin{figure}[t]
  \centering
  \includegraphics[width=0.95\columnwidth]{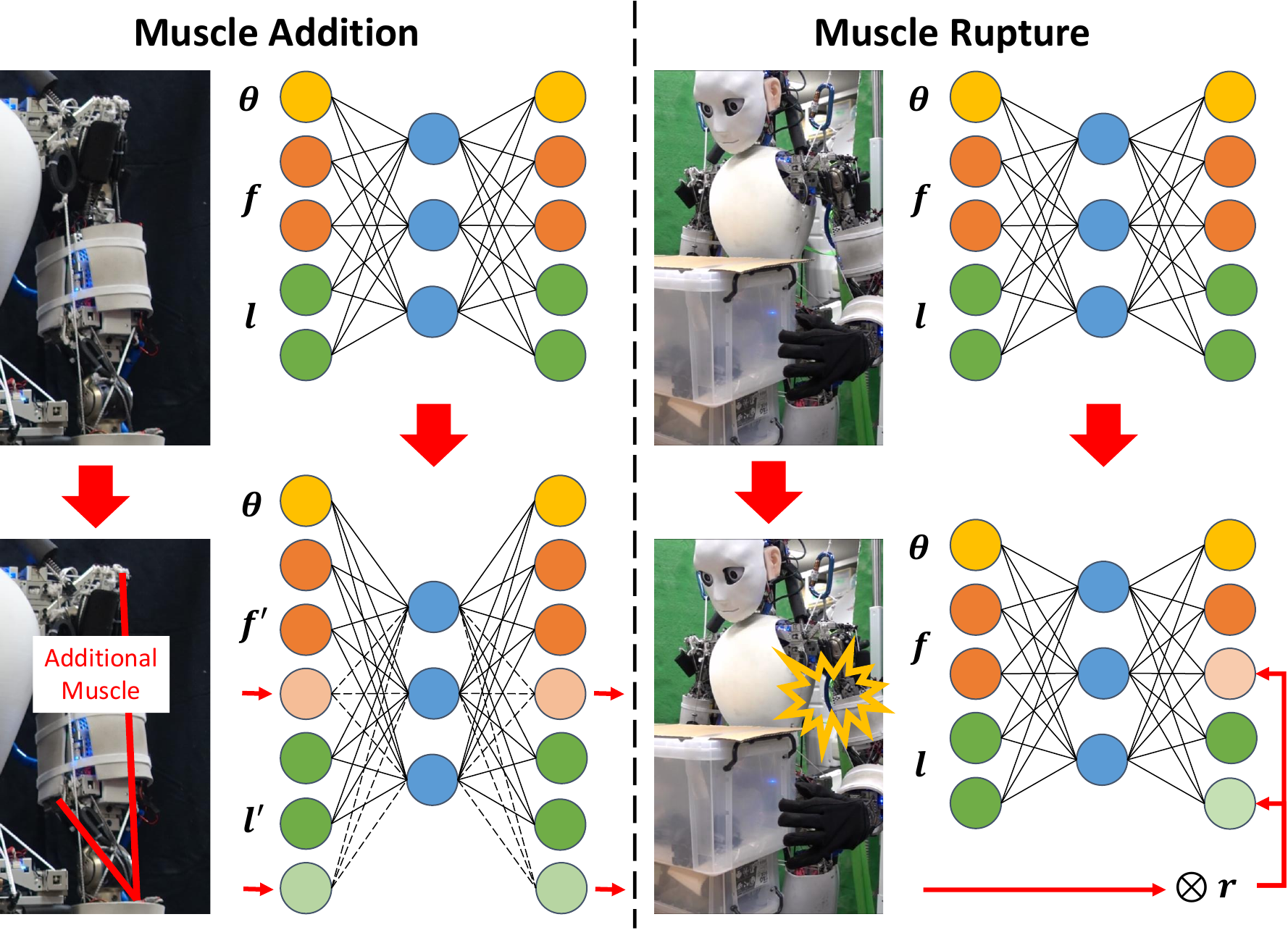}
  \caption{The adaptation for the change in body schema due to muscle addition and muscle rupture.}
  \label{figure:adaptation}
  \vspace{-1.0ex}
\end{figure}

\subsection{Body Schema Adaptation for Musculoskeletal Humanoids} \label{subsec:adaptation}
\switchlanguage%
{%
  In musculoskeletal humanoids, the body schema can change due to various factors.
  Due to Redundancy and Independency of muscles, musculoskeletal humanoids can adapt their body schema in response to bodily changes such as muscle addition \cite{kawaharazuka2022additional} and muscle rupture \cite{kawaharazuka2022redundancy}, as shown in \figref{figure:adaptation}.

  In muscle addition, new muscles are added to the musculoskeletal humanoid according to the task, and the body schema is relearned.
  If a task requires more force, muscles can be added to modify the body to provide sufficient force.
  In this case, the static body schema will have an increased dimension for muscle tension and muscle length.
  Therefore, a new static body schema corresponding to the new number of muscles is constructed, and the weights of the original body schema are copied.
  Subsequently, the network's weights for the new muscles are relearned using random movements while ensuring that the relationships between the original muscles and joints are preserved.

  In muscle rupture adaptation, if a muscle in the musculoskeletal humanoid ruptures for some reason, the body schema is updated to allow the remaining muscles to continue the task.
  In this case, the muscle rupture is detected through the prediction error of the body schema, and a mask variable $\bm{r}$ is introduced to mask the ruptured muscle in the body schema.
  For the original network training, joint angle estimation, posture control, etc., network inputs and outputs are partially masked by $\bm{r}$ to ignore the ruptured muscles.
  This enables control, state estimation, and training to continue without being affected by the ruptured muscles.
}%
{%
  筋骨格ヒューマノイドにおいては, 様々な要因で身体図式が変化する.
  筋肉の冗長性や独立性から, 筋骨格ヒューマノイドは筋の破断や追加に応じて身体図式を適応させることができる.
  \figref{figure:adaptation}に示すように, ここでは筋追加対応\cite{kawaharazuka2022additional}と筋破断対応\cite{kawaharazuka2022redundancy}についてそれぞれ述べる.

  筋追加対応では, 筋骨格ヒューマノイドにタスクに応じて新しい筋を追加し, 身体図式を再学習させる.
  力の足りないタスクがあれば, それに応じて筋を追加し, 十分な力を持つように身体を変更すれば良い.
  このとき, 静的身体図式は筋張力と筋長の次元が一つ増えることになる.
  そのため, 新しい筋数に対応した静的身体図式を構築し, 元々の身体図式の重みをコピーする.
  そのうえで, これまでの身体図式, つまり元々あった筋と関節同士の関係を崩さないように, ランダムな動きから新しい筋に対するネットワークの重みを再学習する.

  筋破断対応では, 筋骨格ヒューマノイドの筋が何らかの理由で破断したときに, 身体図式を更新することで残りの筋でタスクを継続する.
  この場合, 筋破断は身体図式の予測誤差から計算し, 身体図式において破断した筋をマスクする変数$\bm{r}$を用意する.
  これまでのオンライン学習や関節角度推定, 姿勢制御等について, ネットワーク入力や出力を$\bm{r}$により一部マスクし, 破断した筋を無視することで, 破断した筋の影響を受けずに制御や状態推定・学習を継続することができる.
}%

\begin{figure*}[t!]
  \centering
  \includegraphics[width=1.7\columnwidth]{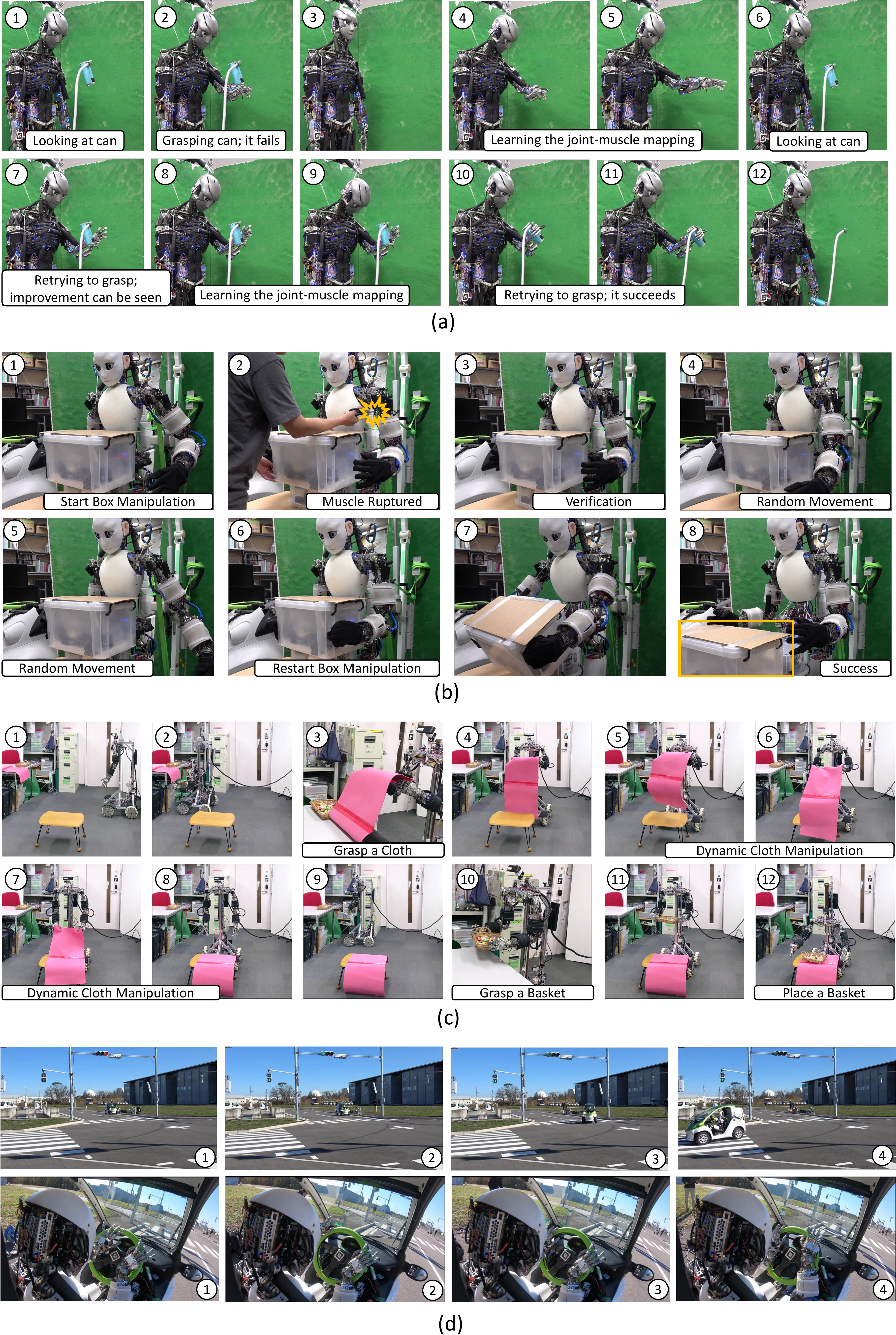}
  \vspace{-1.0ex}
  \caption{Experiments on Kengoro and Musashi: (a) shows the object grasping experiment on Kengoro with online learning of the static body schema, (b) shows the muscle rupture response experiment on Musashi, (c) shows the table setting operation experiment including deformable cloth manipulation by Musashi-W, and (d) shows the integrated autonomous driving experiment by Musashi.}
  \label{figure:experiments}
  \vspace{-2.0ex}
\end{figure*}

\section{Realization of Various Tasks} \label{sec:experiments}
\switchlanguage%
{%
  We present examples that demonstrate the effectiveness of the characteristics of musculoskeletal humanoids and their management and utilization methods through several experiments \cite{kawaharazuka2018online, kawaharazuka2022redundancy, kawaharazuka2022cloth, kawaharazuka2020driving}.
  As this paper is a perspective review, we summarize representative experimental sequences here; detailed quantitative evaluations and method-by-method comparisons are provided in the original papers cited above.

  First, in \figref{figure:experiments} (a), we show an object-grasping experiment based on online learning for the musculoskeletal humanoid Kengoro.
  Here, an accurate body schema is gradually acquired through online learning of the static body schema, allowing object grasping through control based on this schema.
  Initially, in frame \ctext{2}, inverse kinematics on the can fails to achieve the intended grasp.
  However, through online learning in frames \ctext{4}--\ctext{5} and \ctext{8}--\ctext{9}, successful grasping is eventually achieved.

  Next, \figref{figure:experiments} (b) shows a muscle rupture adaptation experiment with the musculoskeletal humanoid Musashi.
  When a muscle rupture occurs, it is detected as an anomaly from the prediction error of the static body schema in frame \ctext{3}.
  The body schema is then relearned to account for the ruptured muscle through random movements in frames \ctext{4}--\ctext{5}, ultimately succeeding in the intended object manipulation.

  Then, \figref{figure:experiments} (c) illustrates a table-setting task involving flexible cloth manipulation using the musculoskeletal wheeled robot Musashi-W.
  Here, muscle relaxation control and muscle thermal control are employed as reflex controls.
  A static body schema is used for regular movements, and a dynamic body schema is used for dynamic flexible cloth manipulation to complete the series of tasks.
  In particular, during dynamic cloth manipulation, adaptively modulating the body stiffness through Variable Stiffness Control improves the maximum joint velocity by 12\%, successfully enabling the cloth to be spread more widely.
  The robot successfully recognizes the tablecloth, grasps it, manipulates it using a dynamic body schema that takes cloth images as input and output, sets it on the table, and then places snacks on the table.

  Finally, \figref{figure:experiments} (d) presents an autonomous driving experiment with the musculoskeletal humanoid Musashi.
  Muscle relaxation control and muscle thermal control are used as reflex controls, with a static body schema employed for steering and a dynamic body schema used for pedal operation.
  The robot successfully recognized traffic signals, operated the pedals using a dynamic body schema with vehicle speed as input and output, and turned the steering wheel to navigate through intersections.
}%
{%
  本研究では, これまで述べた筋骨格ヒューマノイドの特性とその管理・利用方法について, いくつかの実験例を通じてその有効性を示した例を紹介する.
  なお, 本論文はperspective論文であるため, ここでは代表的な実験シーケンスを示すに留め, 定量的な評価や手法間の比較は各引用論文に譲る.

  まず, \figref{figure:experiments}の(a)に, 筋骨格ヒューマノイドKengoroにおける逐次的な学習に基づく物体把持実験を示す\cite{kawaharazuka2018online}
  ここでは, 静的身体図式のオンライン学習により徐々に正確な身体図式を獲得し, 静的身体図式を用いた制御から物体把持を行う.
  はじめ\ctext{2}では缶に対して逆運動学を解いても思ったように把持できないが, \ctext{4}--\ctext{5}や\ctext{8}--\ctext{9}のようにオンラインで再学習を行うことで, 最終的に把持に成功している.

  次に, \figref{figure:experiments}の(b)に, 筋骨格ヒューマノイドMusashiにおける筋破断対応実験を示す\cite{kawaharazuka2022redundancy}
  筋肉が破断したときに, \ctext{3}で静的身体図式の予測誤差からそれを異常検知し, \ctext{4}--\ctext{5}のランダムな動きから筋破断を考慮した上で身体図式を再学習, 最終的に意図した物体操作に成功している.

  次に, \figref{figure:experiments}の(c)に, 筋骨格台車型ロボットMusashi-Wによる柔軟布操作を含むテーブルセッティング動作実験を示す\cite{kawaharazuka2022cloth}.
  ここでは反射制御として筋弛緩制御や筋温度制御, 通常の動作には静的身体図式, 柔軟布操作には動的身体図式を用いることで, 一連のタスクを実現している.
  特に動的柔軟布操作の際には, 身体の剛性を学習で適応的に変化させることで, 関節の最大速度を12\%向上させ, 布を大きく広げることに成功している.
  テーブルクロスを認識して把持し, それを布画像を入出力とした動的身体図式により操作, テーブルの上にセットし, お菓子を把持してテーブルの上に置くことに成功した.

  最後に, \figref{figure:experiments}の(d)に, 筋骨格ヒューマノイドMusashiによる自動運転実験を示す\cite{kawaharazuka2020driving}.
  ここでは反射制御として筋弛緩制御や筋温度制御, ハンドル操作には静的身体図式, ペダル操作には動的身体図式を用いることで, 一連のタスクを実現している.
  信号を認識し, 車体速度を入出力とする動的身体図式によりペダルを操作, ハンドルを切って交差点を曲がることに成功した.
}%

\section{Discussion} \label{sec:discussion}
\switchlanguage%
{%
  In this study, we attempted an integrated explanation by organizing the characteristics of muscles in musculoskeletal humanoids and the advantages and disadvantages arising from them.
  We built a comprehensive system to overcome and leverage these disadvantages and advantages, which includes reflex control, body schema learning, muscle grouping, and body schema adaptation.
  Reflex control encompasses stretch reflex control, antagonist inhibition control, muscle thermal control, maximum speed control, and muscle relaxation control.
  Body schema learning includes static body schemas that represent joint-muscle mapping and dynamic body schemas that represent state equations.
  Muscle grouping involves both manual and automatic grouping, while body schema adaptation includes muscle addition and muscle rupture adaptation.
  This integrated system can manage the shortcomings of musculoskeletal humanoids and make effective use of their advantages, enabling tasks such as object grasping, dynamic flexible object manipulation, and autonomous driving.

  Stretch reflex control, antagonistic inhibition control, and muscle relaxation control have proven useful in certain movements.
  However, when applied to precise manipulations, they can interfere with the accuracy of movements, and are therefore typically disabled.
  These reflexes reduce muscle load and function as a sort of safety mechanism, but as a result, posture may shift slightly.
  Humans also increase body stiffness and move cautiously during precise tasks \cite{gribble2003role}, making reflex control that disrupts these antagonistic relationships unsuitable for precise movements.
  A mechanism to switch reflexes will likely become more important in the future.
  Additionally, muscle thermal control is always effective, while maximum speed control is only effective during rapid movements.

  Currently, the estimation of joint angle in Kengoro relies on muscle length changes and vision.
  However, since this is based on external sensory input, it is not highly reliable; for instance, joint angle cannot be estimated if vision is obstructed.
  In contrast, using static body schema learning gradually enables joint angle estimation from muscle length and muscle tension.
  Internal sensing encompasses not only muscle length and muscle tension but also various types of information, such as IMUs, joint capsule pressure, and skin stretch.
  In the future, we aim to implement these sensors and develop a more accurate and reliable method for estimating the state of the musculoskeletal system.

  This study does not address energy storage and release through redundancy and nonlinear elastic elements.
  Previous research has utilized muscle energy storage and release in exoskeletons \cite{wiggin2011exoskeleton} and jumping spider robots \cite{sprowitz2017spiders}.
  Naturally, this concept is also applicable to musculoskeletal humanoids, but the current challenge lies in nonlinear elastic elements.
  The nonlinear elastic elements used in Musashi \cite{kawaharazuka2019musashi} achieve nonlinear elasticity using rubber O-rings.
  However, the high viscosity of the rubber reduces the effectiveness of energy storage and release.
  On the other hand, nonlinear elastic elements constructed with springs \cite{nakanishi2011kenzoh} become too large to apply to robots like Musashi or Kengoro, which have numerous muscles.
  Future discussions must address the shapes, characteristics, and utilization of nonlinear elasticity.

  In this study, we have not deeply examined robustness against disturbances such as dynamic obstacles or forces exerted by the external environment.
  Control of musculoskeletal humanoids is still in its early stages of development, and current research has mainly focused on internal issues such as Modeling Difficulty, High Internal Force, and High Computational Cost, while stability in interaction with the external environment has not been extensively discussed.
  On the other hand, Variable Stiffness Control through body schema learning, as well as reflex-based controls such as stretch reflex control, muscle thermal control, and muscle relaxation control, are effective against dynamic obstacles and external forces from the environment.
  In future work, we plan to investigate the usefulness of control and learning from a broader perspective that includes not only the internal body dynamics but also interactions with the environment.

  This study does not focus deeply on the hardware aspects.
  The advantages of musculoskeletal humanoids go beyond control and are largely influenced by their physical structures.
  For example, complex joints like scapulae and passive elements like the spine can be created \cite{osada2011planar}, or gear ratios can be adjusted according to the movement \cite{kim2014nonlinear}.
  The flexibility to rearrange components allows for easy reconfiguration.
  On the other hand, this study deliberately avoids deep discussion on hardware, focusing primarily on software, aiming to present a clearer understanding of muscle management and utilization methods.
  Future research will examine hardware aspects in greater detail to achieve long-term, stable, and more human-like movements in musculoskeletal humanoids.

  Currently, discussions from a developmental and cognitive perspective are not included.
  This is because operating a full-body musculoskeletal humanoid requires a significant amount of engineering, making it a time-intensive process to reach a stage where cognitive development can be discussed.
  Body schema learning is highly beneficial from a cognitive development perspective as well.
  Future work will explore its internal states, comparisons with humans, muscle synergies, and related topics.
}%
{%
  本研究では, 筋骨格ヒューマノイドの筋肉の特徴とそこから生じる利点・欠点について整理し, 統合的な説明を試みた.
  利点と欠点を克服・利用するためのシステムとして, 反射制御・身体図式学習・筋グルーピング・身体変化適応を含む全体システムを構築した.
  反射制御は伸長反射制御や拮抗筋抑制制御, 筋温度制御, 最大速度制御, 筋弛緩制御を含み, 身体図式学習には関節-筋空間マッピングを表現する静的身体図式と状態方程式を表現する動的身体図式を含む.
  筋グルーピングには手動グルーピングと自動グルーピング, 身体変化適応には筋追加対応と筋破断対応が含まれた.
  これらを統合したシステムは筋骨格ヒューマノイドの欠点を管理し, 利点を上手に利用することが可能であり, 物体把持や動的柔軟物操作, 自動運転までが可能であった.
  以下にいくつかの議論を示す.

  伸長反射制御や拮抗筋抑制制御, 筋弛緩制御はいくつかの動作において有用であった.
  一方で, 正確なマニピュレーションに利用すると動作の正確性に支障をきたす場合があり通常は切っている.
  これらの反射は筋肉の負荷を減らしある種の安全装置として働くが, それゆえに姿勢が多少変化してしまう.
  人間も正確な動作の際は体の剛性を高めて慎重に動くため\cite{gribble2003role}, その拮抗関係を修正してしまうような反射制御は正確な動作には向かず, 反射を切り替える仕組みも今後重要になると考えている.
  加えて, 筋温度制御は常に有効であり, 最大速度制御は素早い動きをするときのみ有効である.

  現在Kengoroについては関節角度を筋長変化と視覚から推定している.
  これは視覚という外部感覚に頼っているため, あまり信頼できるものではなく, 例えば視覚が遮られた場合には関節角度の推定ができなくなる.
  これに対して, 静的身体図式学習を用いることで, 徐々に関節角度を筋長と筋張力から推定することができるようになる.
  内部感覚は筋長と筋張力だけでなく, IMUや関節包内の圧力, 皮膚の伸びなどの様々な情報が含まれる.
  今後これらセンサを実装しつつ, より正確かつ信頼できる筋骨格系の状態推定を開発する予定である.

  本研究では, 冗長性と非線形弾性要素によるエネルギーの蓄積と解放については扱えていない.
  これまでexoskeletonやjumping spider robotにおいて筋肉のエネルギー蓄積と解放が用いられている\cite{wiggin2011exoskeleton, sprowitz2017spiders}
  もちろん同様の話は筋骨格ヒューマノイドにおいても有効であると考えられるが, 現状の問題は非線形弾性要素にある.
  Musashiに用いられている非線形弾性要素\cite{kawaharazuka2019musashi}はO-ringを活用してゴムにより非線形弾性を得るものである.
  これは確かに非線形弾性を得ることはできるが, ゴムの粘性が高いためエネルギーの蓄積と解放の効果があまり高くない.
  一方で, バネによって構成される非線形弾性要素\cite{nakanishi2011kenzoh}はサイズが大きくなってしまうため, MusashiやKengoroのように多数の筋肉を有するロボットには適用できない.
  今後非線形弾性の形状や特性, その活用方法についても議論を進めていく必要がある.

  本研究では, 動的な障害物や外部環境から受ける力などの外乱に対する頑健性については深く考察できていない.
  筋骨格ヒューマノイドの制御はまだ発展途上であり, 現状は主に身体内部のModeling DifficultyやHigh Internal Force, High Computational Constに焦点が当たっており, 外部環境とのインタラクションにおける安定性については大きく議論されていない.
  一方で, 身体図式学習におけるVariable Stiffness Controlや, 反射制御におけるStretch Reflex Control (SRC), Muscle Thermal Control (MTC), Muscle Relaxation Control (MRC)などは動的障害物や環境からの外力に対して効果を発揮する.
  今後, 身体内部のみでなく, 環境を含むより広い視点から制御・学習の有用性について考察していく予定である.

  本研究ではハードウェアについては深く触れていない.
  筋骨格ヒューマノイドにおける利点は制御だけに留まらず, その身体構造によるところが大きい.
  例えば, 肩甲骨のような複雑な関節や背骨のような受動要素を作ることができたり\cite{osada2011planar}, ギア比を動きに合わせて変化させることができたりする\cite{kim2014nonlinear}.
  比較的自由に設計変更ができるため, 構成・再構成が容易な点も大きい.
  一方で, 本研究は敢えてハードウェアについて深く触れずに, あくまでソフトウェアに大きく焦点を当てた内容である.
  今後は, これらの点についてもより詳細に議論し, 筋骨格ヒューマノイドの長期的で安定したより人間らしい動作を実現するための研究を進めていく予定である.

  現状発達認知の観点からの議論は出来ていない.
  これは, 全身筋骨格ヒューマノイドを動かすには非常に多くのエンジニアリングが必要であり, 認知発達等について議論するような段階に到達するために多大な時間がかかるためである.
  身体図式学習については認知発達の観点からも非常に有益であり, その内部状態の解析や人間との比較, 筋シナジー等の議論は今後展開していく予定です.
}%

\section{Conclusion} \label{sec:conclusion}
\switchlanguage%
{%
  In this study, we organized the characteristics of muscles in musculoskeletal humanoids and the advantages and disadvantages that arise from them.
  Using the musculoskeletal humanoids Kengoro and Musashi as examples, we conducted an integrated discussion on how to empirically manage their disadvantages and leverage their advantages.
  The characteristics of muscles include Redundancy, Independency, Anisotropy, Variable Moment Arm, and Nonlinear Elasticity.
  From these characteristics arise advantages and disadvantages such as Modeling Difficulty, High Internal Force, High Computational Cost, Easy Muscle Arrangement, Variable Stiffness Control, Muscle Rupture Handling, Nonlinearity Utilization, and Free Joint Utilization.
  These can be managed and utilized through low-level reflex control, high-level learning control, muscle grouping, and body schema adaptation in musculoskeletal humanoids.
  Several experiments have shown that long-term and stable movements can be realized by musculoskeletal humanoids through these methods.
  We hope this study provides a unified discussion on the characteristics of musculoskeletal structures and their management and utilization methods, serving as a foundation for future research.
}%
{%
  本研究では, 筋骨格ヒューマノイドの筋肉の特徴とそこから生じる利点・欠点について整理した.
  また, これまで我々が開発してきた筋骨格ヒューマノイドKengoroとMusashiを例に, その欠点をどのように管理し, 利点をどのように利用できるのかについて統合的な議論を行った.
  筋肉の特徴には冗長性・独立性・異方性・可変モーメンタム・非線形弾性があり, これらから, モデル化困難性・拮抗関係による内力発生・計算量増大・容易な筋配置・可変剛性制御・筋破断対応・非線形性の利用・フリー関節の利用という利点・欠点が生じる.
  それらは筋骨格ヒューマノイドにおける低レイヤの反射制御と高レイヤの学習制御, それに伴う筋のグルーピングと身体変化への適応によって管理・利用可能である.
  これらの研究により, 長期的で安定した動作が筋骨格ヒューマノイドによって実現可能であることがいくつかの実験から示されている.
  本研究が, 筋骨格構造の特性とその管理・利用方法についての統一的な議論を提供し, 今後の研究に対する基盤となることを期待する.
}%

\section*{Acknowledgments}
\switchlanguage%
{%
  This work was partially supported by JST FOREST Grant Number JPMJFR232N, Japan.
}%
{%
  本研究はJST FOREST課題番号JPMJFR232Nの支援を受けて行われた.
}%

{
  \bibliographystyle{IEEEtran}
  \bibliography{main}
}

\end{document}